\documentclass[11pt]{article}
\usepackage{graphicx}
\usepackage{color}
\usepackage{amsmath}
\usepackage{amssymb}
\usepackage{amscd}
\usepackage{bbm}
\newcommand{\R}{\mathbb{R}}
\newcommand{\inr}[1]{\bigl< #1 \bigr>}

\newcommand{\E}{\mathbb{E}}

\newcommand{\eps}{\varepsilon}

\newtheorem{Theorem}{Theorem}[section]
\newtheorem{Lemma}[Theorem]{Lemma}

\newtheorem{Definition}[Theorem]{Definition}
\newtheorem{Problem}[Theorem]{Problem}

\newtheorem{Corollary}[Theorem]{Corollary}
\newtheorem{Remark}[Theorem]{Remark}
\newtheorem{Example}[Theorem]{Example}
\newtheorem{Assumption}{Assumption}[section]

\numberwithin{equation}{section}

\def \proof {\noindent {\bf Proof.}\ \ }

\def \endproof
{{\mbox{}\nolinebreak\hfill\rule{2mm}{2mm}\par\medbreak}}
\def\IND{\mathbbm{1}}

\begin{document}
\title{\bf Learning without Concentration for General Loss Functions
}

\author{
\\{\bf Shahar Mendelson}
\thanks{Department of Mathematics, Technion -- Israel Institute of Technology, email: shahar@tx.technion.ac.il. Partially supported by the Mathematical Sciences Institute -- The Australian National University and by ISF grant 900/10.}
}

\maketitle

\section{Introduction}
Prediction and estimation problems play a major role in modern mathematical statistics. The aim is to approximate, in one way or another, an unknown random variable $Y$ by a function from a given class $F$, defined on a probability space $(\Omega,\mu)$. The given data is a random sample $(X_i,Y_i)_{i=1}^N$, distributed according to the $N$-product of the joint distribution of $\mu$ and $Y$, endowed on the product space $(\Omega \times \R)^N$.

The notion of approximation may change from problem to problem. It is reflected by different choices of {\it loss functions}, which put a price tag on predicting $f(X)$ instead of $Y$. Although it is not the most general form possible, we will assume throughout this article that if $\ell$ is the loss function, the cost of predicting $f(X)$ instead of $Y$ is $\ell(f(X)-Y)$. Formally,

\begin{Definition} \label{def:loss-functional}
A loss is a real-valued function that is even, increasing in $\R_+$ and convex, and vanishes in $0$. We will assume that it is sufficiently smooth -- for example, that it has a second derivative, except, perhaps in $\pm x_0$ for some fixed $x_0$ -- although, as will be clear from what follows, this assumption can be relaxed further.
\end{Definition}

Once the loss is selected, one can define the best element in the class, namely, a function in $F$ that minimizes the average loss, or {\it risk}, $\E \ell(f(X)-Y)$ (with the obvious underlying assumption that the minimizer exists). We will also assume that the minimizer, denoted by $f^*$, is unique, though this assumption can be relaxed.

Next, one may choose a procedure that uses the data $(X_i,Y_i)_{i=1}^N$ to produce a (random) function $\hat{f} \in F$.

The effectiveness of $\hat{f}$ may be measured in several ways, and the two we will focus on here lead to the {\it prediction/estimation problem}.

\begin{Problem} \label{qu:main}
Given a procedure $\hat{f}$, find the `smallest' functions ${\cal E}_p$ and ${\cal E}_{e}$ possible for which the following holds. If $F \subset L_2(\mu)$ is a class of functions and $Y$ is the unknown target, then with probability at least $1-\delta$ over samples $(X_i,Y_i)_{i=1}^N$,
$$
\E \left(\ell(\hat{f}(X)-Y)|(X_i,Y_i)_{i=1}^N\right) \leq \inf_{f \in F} \E \ell(f(X)-Y) + {\cal E}_p.
$$

Alternatively, with probability at least $1-\delta$,
$$
\|\hat{f}-f^*\|_{L_2}^2 = \E \left( (\hat{f}-f^*)^2(X) | (X_i,Y_i)_{i=1}^N \right) \leq {\cal E}_e.
$$
The functions ${\cal E}_p$ and ${\cal E}_e$ may depend on the structure of $F$, the sample size $N$, the probability $\delta$, some `global' properties of $Y$ (e.g., its $L_q$ norm), etc.
\end{Problem}

${\cal E}_p$ measures the `predictive capabilities' of $\hat{f}$, that is, whether $\hat{f}$ is likely to be almost as effective as the best possible in the class - $f^*$. ${\cal E}_e$ measures the distance between $\hat{f}$ and $f^*$, with respect to the underlying $L_2(\mu)$ metric.

\vskip0.5cm

The amount of literature centred around the theory of prediction and estimation is extensive and goes well beyond what can be reasonably surveyed here. We refer the reader to the manuscripts \cite{MR1240719}, \cite{DGL:96}, \cite{vanderVaartWellner}, \cite{MR2319879}, \cite{MR2829871}, \cite{MR2724359} and \cite{MR2807761} as possible starting points for information on the history of the problem as well as for more recent progress.
\vskip0.5cm
The procedure we will focus on here is empirical risk minimization (ERM), in which ${\hat f}$ is selected to be a function in $F$ that minimizes the empirical risk
$$
P_N \ell_f \equiv \frac{1}{N} \sum_{i=1}^N \ell(f(X_i)-Y_i),
$$
where here, and throughout the article, $P_N$ denotes the empirical mean associated with the random sample.

Since it is impossible to obtain nontrivial information on the performance of {\it any} procedure, including ERM, without imposing some assumptions on the class $F$, the target $Y$ and the loss $\ell$, one has to select a framework that, on one hand, is general enough to include natural problems that one would like to study, but on the other, still allows one to derive significant results on prediction and estimation.

Unfortunately, some of the assumptions that are commonly used in literature are highly restrictive, though seemingly benign. And, among the more harmful assumptions are that the loss is a Lipschitz function and that functions in $F$ and $Y$ are uniformly bounded.

The origin of these assumptions is technical: they are an outcome of the `classical' method of analysis used to tackle Problem \ref{qu:main}. The method itself is based on tools from Empirical Processes Theory, most notably, on contraction and concentration arguments that are simply false without imposing the right assumptions on the class, the target and the loss. However, the assumptions leave a large number of natural problems out of reach.

We will present an example of the `classical' method in Appendix \ref{sec:the-classical-method} in some detail, but for the time being, let us present an outline of its main ideas and shortcomings.

The basic underlying assumption behind data-driven procedures like ERM is that sampling mimics reality. Since one's goal is to identify $f^*$ -- a function that minimizes in $F$ the functional $f \to \E \ell(f(X)-Y)$, a natural course of action is to compare empirical means of the loss functional to the actual means.

To that end, consider the excess loss functional associated with $f \in F$,
$$
{\cal L}_f(X,Y) = \ell(f(X)-Y)-\ell(f^*(X)-Y).
$$
Observe that for every $f \in F$, $\E {\cal L}_f \geq 0$, and if $f^*$ is unique equality is achieved only by $f^*$. On the other hand, since ${\cal L}_{f^*}=0$, it is evident that $P_N {\cal L}_{\hat f} \leq 0$; thus, for every sample, the empirical minimizer belongs to the random set
$$
\{f \in F : P_N {\cal L}_f \leq 0\}.
$$
The key point in the analysis of ERM is that the random set of potential minimizers consists of functions for which sampling behaves in an a-typical manner: $P_N {\cal L}_f \leq 0$ while $\E {\cal L}_f > 0$. Thus, one may identify the set by building on the discrepancy between the `empirical' and `actual' behaviour of means. For example, a solution to the prediction problem follows if this set consists only of functions with `predictive capabilities' that are close to the optimal in $F$, while the estimation problem may be resolved if the random set consists only of functions that are close to $f^*$ with respect to the $L_2$ distance.

\vskip0.5cm

What makes the nature of the set $\{f : P_N {\cal L}_f \leq 0\}$ rather elusive is not only the fact that it is random, but also that one has no real knowledge of the functions $\ell_f=\ell(f(X)-Y)$ and ${\cal L}_f=\ell_f -\ell_{f^*}$, as the two have unknown components - the target $Y$ and the function $f^*$.

The core idea in the `classical method' is to identify $\{f \in F : P_N{\cal L}_f \leq 0\}$ by applying concentration results for a well-chosen subset of excess loss functions $\{{\cal L}_f : f \in F^\prime\}$, thus showing that for a large subset $F^\prime \subset F$, $P_N {\cal L}_f$ cannot be too far from $\E{\cal L}_f$ (or, for more sophisticated results, that the ratios $P_N {\cal L}_f/\E {\cal L}_f$ cannot be too far from $1$). Since $\E {\cal L}_f > 0$ if $f \not = f^*$, this forces $f \to P_N {\cal L}_f$ to be positive on $F^\prime$ and thus $\hat{f} \in F \backslash F^\prime$.

Naturally, concentration results come at a cost, and estimates such as
\begin{equation} \label{eq:concentration-intro}
\sup_{f \in F^\prime} \left|P_N{\cal L}_f - \E {\cal L}_f\right| < \eps \ \ {\rm or} \ \ \sup_{f \in F^\prime} \left|\frac{P_N {\cal L}_f}{\E {\cal L}_f}-1\right| < \eps
\end{equation}
require strong assumptions on the random variables involved -- for example, that functions in $F$ and $Y$ are uniformly bounded (see the books \cite{Ledoux:2001,BouLugMass13} for more details on concentration of measure phenomena).

\vskip0.5cm

The need for two-sided concentration estimates has been the driving force behind the assumption that functions in $F$ and $Y$ are uniformly bounded. And, although one can relax the uniform boundedness assumption (see, e.g., \cite{LM13}) and still obtain \eqref{eq:concentration-intro}, a necessary condition for two-sided inequalities like \eqref{eq:concentration-intro} is that class members exhibit rapidly decaying tails (e.g. a subgaussian behaviour), still forcing one to impose strong tail assumptions.

Finally, and possibly the most costly step in the classical method is contraction, in which one combines the fact that class members and the target are uniformly bounded functions and that the loss is Lipschitz on the ranges of the functions $f(X)-Y$. This combination allows one to bound the empirical process indexed by the excess loss class using an empirical process indexed by functions of the form $f-f^*$ (see Appendix \ref{sec:the-classical-method} for more details).

\vskip0.5cm

One result that is based on the classical method and that uses the full strength of the two assumptions -- that class members and the target are uniformly bounded and that the loss is Lipschitz, is Theorem \ref{thm:BBM} below, proved originally in \cite{MR2166554}. It will serve as a preliminary benchmark for our discussion.

Assume that $F$ is a class of functions that are bounded by $1$. Let $\ell$ be a Lipschitz function with constant $\|\ell\|_{{\rm lip}}$ on $[-2,2]$, which is an interval containing all the ranges of $f(X)-Y$ for the unknown target $Y$ that is also bounded by $1$. Assume further that $f^*$ exists and is unique and that for every $f \in F$, $\|f-f^*\|_{L_2}^2 \leq B \E {\cal L}_f$, which is the more significant part of the so-called Bernstein condition (see, e.g., \cite{lbw:96} and \cite{MR1930004,MR2426759}).

One example in which all these conditions hold is when $F$ is a closed, convex class consisting of functions into $[-1,1]$, as is $Y$, and $\ell(t)=t^2$. Hence, it follows that $B=1$ and $\|\ell\|_{\rm lip}=4$ (for more information see Appendix \ref{sec:the-classical-method} and \cite{MR2166554}).

\vskip0.5cm

Let $D_{f^*}$ be the $L_2(\mu)$ ball of radius $1$, centred in $f^*$. Thus,
$\{f \in F : \|f-f^*\|_{L_2} \leq r\} = F \cap rD_{f^*}$.
For every $r>0$, let
\begin{equation} \label{eq:k-N}
k_N(r)=\sup_{f \in F \cap rD_{f^*}} \left|\frac{1}{\sqrt{N}}\sum_{i=1}^N\eps_i (f-f^*)(X_i)\right|,
\end{equation}
and
\begin{equation} \label{eq:bar-k-N}
\bar{k}_N(r)=\E\sup_{f \in F \cap rD_{f^*}} \left|\frac{1}{\sqrt{N}}\sum_{i=1}^N\eps_i (f-f^*)(X_i)\right|,
\end{equation}
where $(\eps_i)_{i=1}^N$ are independent, symmetric, $\{-1,1\}$-valued random variables that are independent of $(X_i)_{i=1}^N$, and the expectation is taken with respect to both $(X_i)_{i=1}^N$ and $(\eps_i)_{i=1}^N$. Finally, set
$$
k_N^*(\gamma,\delta)=\inf \left\{r>0: Pr \left(k_N(r/\|\ell\|_{{\rm lip}})\leq \gamma r^2\sqrt{N}\right) \geq 1-\delta \right\}
$$
and
$$
\bar{k}_N^*(\gamma)=\inf \left\{r>0: \bar{k}_N(r/\|\ell\|_{{\rm lip}})\leq \gamma r^2\sqrt{N}\right\}.
$$

\begin{Theorem} \label{thm:BBM}
There exist absolute constants $c_1$ and $c_2$ for which the following holds.
If $F$, $Y$ and $\ell$ are as above, then for every $0<\delta<1$, with probability at least $1-\delta$
\begin{equation}
  \label{eq:BBM-prediction-delta}
\E{\cal L}_{\hat f} \leq c_1\max\left\{\left(k_N^*\left(c_2\left(B\|\ell\|_{{\rm lip}}\right)^{-1},\delta\right)\right)^2,\frac{\|\ell\|_{{\rm lip}}^2B}{N}\right\},
\end{equation}
and
\begin{equation}
  \label{eq:BBM-prediction}
\E{\cal L}_{\hat f} \leq c_1\max\left\{\left(\bar{k}_N^*\left(c_2(B\|\ell\|_{{\rm lip}})^{-1}\right)\right)^2,(\|\ell\|_{{\rm lip}}^2B)\frac{\log(1/\delta)}{N}\right\}.
\end{equation}
\end{Theorem}
By the Bernstein condition, $\|f-f^*\|_{L_2}^2 \leq B \E {\cal L}_f$ for every $f \in F$, and thus analogous results hold for the estimation problem.
\vskip0.5cm

A version of \eqref{eq:BBM-prediction} will be presented in Appendix \ref{sec:the-classical-method}.
\vskip0.5cm
It is not difficult to see that the assumptions involved in Theorem \ref{thm:BBM} are rather restrictive. For example, Theorem \ref{thm:BBM} cannot be used to tackle one of the most fundamental problems in Statistics -- linear regression in $\R^n$ relative to the squared loss and with  independent gaussian noise.

\begin{Example} \label{ex:regression-gaussian-noise}
Let $\ell(x)=x^2$. Given $T \subset \R^n$, set $F_T =\left\{\inr{t,\cdot} : t \in T\right\}$ to be the class of linear functionals on $\R^n$ associated with $T$. Let $\mu$ be a measure on $\R^n$, set $X$ to be a random vector distributed according to $\mu$ and put $W$ to be a standard gaussian variable that is independent of $X$. The target is given by $Y=\inr{t^*,\cdot}+W$ for some fixed but unknown $t^* \in T$.

Observe that
\begin{description}
\item{$\bullet$} $Y$ is not bounded (because of the gaussian noise).
\item{$\bullet$ } Unless $\mu$ is supported in a bounded set in $\R^n$, functions in $F_T$ are not bounded.
\item{$\bullet$}  The loss $\ell(x)=x^2$ satisfies a Lipschitz condition in $[-a,a]$ with a constant $2a$. Unless $\mu$ has a bounded support and $Y$ is bounded, $\ell$ does not satisfy a Lipschitz condition on an interval containing the ranges of the functions $f(X)-Y$.
\end{description}
Each one of these observations is enough to force linear regression with a gaussian noise outside the scope of Theorem \ref{thm:BBM}. And, what is equally alarming is that the same holds even if $\mu$ is the standard gaussian measure on $\R^n$, regardless of $T$, the choice of noise or even its existence.
\end{Example}

An additional downside of Theorem \ref{thm:BBM} is that even in situations that do fall within its scope, resulting bounds are often less than satisfactory.

One example (out of many) indicating the suboptimal nature of Theorem \ref{thm:BBM} is the {\it persistence problem}, which will be presented in Appendix \ref{app:persistence}.

\vskip0.5cm

The suboptimal behaviour of Theorem \ref{thm:BBM} goes well beyond an isolated example. It is endemic and is caused by the nature of the complexity parameter used to govern the rates ${\cal E}_p$ and ${\cal E}_e$.

Indeed, when considering likely sources of error in prediction or estimation, two generic reasons spring to mind:
\begin{description}
\item{$\bullet$} $(X_1,..,X_N)$ is merely a sample and two functions in $F$ can agree on that sample, but still be very different. This leads to the notion of the {\it version space}: a random subset of $F$, defined by
$$
\{f \in F: f(X_i)=f^*(X_i) \ {\rm for \ every \ } 1 \leq i \leq N\}
$$
and measures the way in which a random sample can be used to distinguish between class members. Clearly, the $L_2(\mu)$ diameter of the version space is an intrinsic property of the class $F$ and has nothing to do with the noise\footnote{We will refer to $f^*(X)-Y$ as the noise of the problem. This name makes perfect sense when $Y=f_0(X)+W$ for a symmetric random variable $W$ that is independent of $X$, and we will use it even when the target does not have that particular form.} $\xi=f^*(X)-Y$. Standard arguments show (see, e.g. \cite{LM13}) that even in noiseless problems, when $Y=f_0(X)$ for some $f_0 \in F$, it is impossible to construct a procedure whose error rates constantly outperform the $L_2$ diameter of the version space.

\item{$\bullet$} Measurements are noisy: one does not observe $f^*(X_i)$, but rather $Y_i$. Since results in certain specific cases, as well as common sense, indicate that the `closer' $Y$ is to $F$, the better the behaviour of ${\cal E}_p$ and ${\cal E}_e$ should be,  ${\cal E}_p$ and ${\cal E}_e$ should depend, in one way or another, on the `noise level' of the problem, as captured by a natural distance between the target and the class.
\end{description}

With this in mind, it is reasonable to conjecture that ${\cal E}_p$ and ${\cal E}_e$ should
exhibit two regimes, captured by two different complexity parameters. Firstly, a `low noise' regime, in which the `noise' $\xi=f^*(X)-Y$ is sufficiently close to zero in the right sense, and the behaviour of ERM is similar to its behaviour in the noise-free problem -- essentially the $L_2$ diameter of the version space. Secondly, a `high noise' regime, in which mistakes occur because of the way the loss affects the interaction between class members and the noise.
\vskip0.5cm

Theorem \ref{thm:BBM} yields only one regime that is governed by a single complexity parameter. This parameter does not depend on the noise $\xi=f^*(X)-Y$, except via a trivial $L_\infty$ bound, and depends solely on the correlation of the set $\{(f(X_i))_{i=1}^N : f \in F\}$ (the so-called random coordinate projection of $F$) with a generic random noise model, represented by a random point in $\{-1,1\}^N$ that may have nothing to do with the actual noise.

\vskip0.5cm

The main goal of this article is to address Problem \ref{qu:main} by showing that ${\cal E}_p$ and ${\cal E}_e$ indeed have two regimes. Each one of those regimes is captured by a different parameter: firstly, an `intrinsic parameter' that depends only on the class and not on the target or on the loss, and which governs low-noise problems, in which $Y$ is sufficiently close to $F$; secondly, an external parameter that captures the interaction of the class with the noise and with the loss, and dominates in high-noise situations, when $Y$ is far from $F$.

Moreover, a solution to Problem \ref{qu:main} has to hold without the restrictive assumptions of Theorem \ref{thm:BBM}, namely:
\begin{description}
\item{$\bullet$} The class $F$ need not be bounded in $L_\infty$, but rather satisfies weaker tail conditions.
\item{$\bullet$} The target $Y$ need not be bounded (in fact, $Y \in L_2$ suffices in most cases).
\item{$\bullet$} The loss function $\ell$ need not be Lipschitz on an interval containing the ranges of $f(X)-Y$.
\end{description}


\subsection{Possible complexity parameters in subgaussian learning} \label{sec:squared-subgaussian}
The two noise regimes and the fact that they are captured by an intrinsic parameter in low-noise situations, and an external parameter in high noise cases was first observed in \cite{LM13} for the problem of subgaussian learning relative to the squared loss. We will sketch that argument here, as it will serve as a more useful benchmark than Theorem \ref{thm:BBM} in what follows. Also, for the sake of brevity, we will only study the estimation problem, as the prediction problem requires an additional argument (see the presentation in \cite{LM13} for more details).

\vskip0.5cm

Observe that if $\ell(t)=t^2$ then for every $f \in F$ and every $(X,Y)$,
$$
{\cal L}_f(X,Y)=(f(X)-Y)^2-(f^*(X)-Y)^2= (f-f^*)^2(X)-2\xi(f-f^*)(X),
$$
where here, and throughout this article, $\xi=\xi(X,Y)=f^*(X)-Y$. Therefore, the centred empirical excess loss process $f \to P_N {\cal L}_f - \E {\cal L}_f$ is a sum of a quadratic term,
\begin{equation} \label{eq:quadratic}
f \to \frac{1}{N} \sum_{i=1}^N (f-f^*)^2(X_i) - \E (f-f^*)^2
\end{equation}
and a multiplier one,
\begin{equation} \label{eq:linear}
f \to \frac{1}{N} \sum_{i=1}^N \xi_i(f-f^*)(X_i) - \E \xi (f-f^*)
\end{equation}
where $\xi_i=f^*(X)-Y$.

By the Gin\'{e}-Zinn symmetrization theorem \cite{MR757767}, the latter is essentially equivalent to the symmetrized multiplier process
\begin{equation} \label{eq:multiplier}
f \to \frac{1}{N} \sum_{i=1}^N \eps_i \xi_i (f-f^*)(X_i).
\end{equation}
Assume that on an event with high probability, one has:
\begin{description}
\item{$\bullet$} If $\|f-f^*\|_{L_2} > r_Q$ then
\begin{equation} \label{eq:quad-subgaussian-cond}
\frac{1}{N}\sum_{i=1}^N (f-f^*)^2(X_i) \geq \frac{1}{2}\|f-f^*\|_{L_2}^2.
\end{equation}
\item{$\bullet$} If $\|f-f^*\|_{L_2} > r_M$ then
\begin{equation} \label{eq:multi-subgaussian-cond}
\left|\frac{1}{N} \sum_{i=1}^N \xi_i (f-f^*)(X_i) - \E \xi (f-f^*)(X) \right| < \frac{1}{4}\|f-f^*\|_{L_2}^2
\end{equation}
(which is equivalent to a similar inequality for the symmetrized process in \eqref{eq:multiplier}).
\end{description}
If, in addition, for every $f \in F$, $\E \xi(f-f^*)(X) \geq 0$, then on the event in question,
\begin{align*}
P_N {\cal L}_f & \geq \frac{1}{N}\sum_{i=1}^N (f-f^*)^2(X_i) - 2 \left|\frac{1}{N} \sum_{i=1}^N \xi_i (f-f^*)(X_i) - \E \xi (f-f^*)(X) \right|
\\
& + 2\E \xi(f-f^*)(X) >\frac{1}{2}\|f-f^*\|_{L_2}^2 - \frac{2}{4}\|f-f^*\|_{L_2}^2>0.
\end{align*}
Hence, on that event, if $\|f-f^*\|_{L_2} \geq \max\{r_Q,r_M\}$ then $P_N {\cal L}_f > 0$ and $f$ is not an empirical minimizer. Therefore,
$$
\|\hat{f}-f^*\|_{L_2} \leq \max\{r_M,r_Q\}.
$$

This decomposition is at the heart of the argument used in \cite{LM13}, under the assumption that $F$ is a convex, $L$-subgaussian class of functions:
\begin{Definition} \label{def:subgaussian}
The $\psi_2$ norm of a function $f$ is
$$
\|f\|_{\psi_2} = \inf \left\{c>0 : \E \exp(|f/c|^2) \leq 2 \right\},
$$
and $f \in L_{\psi_2}$ if $\|f\|_{\psi_2}<\infty$.

A class of functions is $L$-subgaussian if for every $f,h \in F \cup \{0\}$, $\|f-h\|_{\psi_2} \leq L \|f-h\|_{L_2}$.
\end{Definition}

To formulate the result from \cite{LM13} and, in particular, identify in the subgaussian case the parameters $r_Q$, $r_M$ and the high probability event, one requires several additional definitions.

Let $d_F(L_2)$ be the diameter of $F$ in $L_2(\mu)$ and set $\{G_f : f \in F\}$ to be the canonical gaussian process indexed by $F$ with a covariance structure endowed by $L_2(\mu)$. Given a set $F^\prime \subset F$, denote by $\E\|G\|_{F^\prime}$ the expectation of the supremum of $\{G_f : f \in F^\prime\}$ (and throughout the article we will avoid any measurability questions).

Set
$$
k_F=\left(\frac{\E \|G\|_F}{d_F(L_2)}\right)^2,
$$
which is an extension of the celebrated {\it Dvoretzky-Milman dimension} of a convex body in $\R^n$. We refer the reader to \cite{MR856576,MR1036275} for more details on the Dvoretzky-Milman dimension and its role in Asymptotic Geometric Analysis, and to \cite{shahar-psi2} for information on the way $k_F$ captures the structure of $P_\sigma F = \{(f(X_i))_{i=1}^N : f \in F\})$ -- a typical coordinate projection of $F$.

It turns out that one may identify $r_Q$ and $r_M$ using the gaussian parameters $s_Q$ and $s_M$ defined below. For the sake of simplicity, we will assume that $F$ is centrally-symmetric (that is, if $f \in F$ then $-f \in F$), though the modifications needed in the definition when it is not are minor -- as the symmetry allows one to use a ball centred in $0$ rather than in $f^*$.
\begin{Definition} \label{def:fixed-point}
Let $D$ be the unit ball in $L_2(\mu)$. For every $\eta_1,\eta_2>0$, let
\begin{equation}\label{eq:fixed-point-sub-gauss}
s_M(\eta_1)=\inf \left\{ 0<s \leq d_F(L_2) : \E\|G\|_{F \cap s D} \leq \eta_1 s^2 \sqrt{N} \right\},
\end{equation}
and put
$$
s_Q(\eta_2)=\inf \left\{ 0<s \leq d_F(L_2) : \E\|G\|_{F \cap s D } \leq \eta_2 s \sqrt{N} \right\}.
$$
In both cases, if the set is empty, set $s_M(\eta_1)=d_{F}(L_2)$ (resp. $s_Q(\eta_2)=d_{F}(L_2)$).
\end{Definition}

The key feature of the subgaussian setup is that the quadratic and multiplier processes exhibit a strong concentration phenomenon:

\begin{Theorem} \label{thm:subgaussian-quad-and-multi} \cite{shahar-psi2}
There exists absolute constants $c_1$ and $c_3$, and a constant $c_2$ that depends only on $L$ for which the following holds.

Assume that $F$ is an $L$-subgaussian class of functions and that $\xi \in L_{\psi_2}$. For any $t \geq c_1$, with probability at least $1-2\exp(-c_2(L)t^2k_F)$,
\begin{equation} \label{eq:subgaussian-multiplier-process-est}
\sup_{f \in F} \left|\sum_{i=1}^N (\xi_i f(X_i)-\E \xi f ) \right|
\leq c_3 L t^2 \sqrt{N} \|\xi\|_{\psi_2} \E\|G\|_{F},
\end{equation}
and
\begin{equation} \label{eq:subgaussian-quadratic-process-est}
\sup_{f \in F} \left|\sum_{i=1}^N (f^2(X_i)-\E f^2) \right|
\leq c_3 L^2 t^2 \left((\E\|G\|_F)^2 + t\sqrt{N}d_F(L_2) \E\|G\|_{F}\right).
\end{equation}
\end{Theorem}

The fixed point $s_M$ arises from the symmetrized multiplier process. Indeed, by the first part of Theorem \ref{thm:subgaussian-quad-and-multi}, if $H$ is an $L$-subgaussian class and $\xi \in L_{\psi_2}$, then with high probability,
\begin{equation} \label{eq:multi-subgaussian}
\sup_{h \in H} \left|\frac{1}{N} \sum_{i=1}^N \eps_i \xi_i h(X_i)\right| \leq
c(L) \|\xi\|_{\psi_2} \frac{\E\|G\|_{H}}{\sqrt{N}}.
\end{equation}
Note that if $F$ is a convex, centrally symmetric class then
$$
H_s=\{f-h : f,h \in F, \ \|f-h\|_{L_2} \leq s\} \subset 2F \cap sD.
$$
Thus, $s_M$ is chosen to ensure that with high probability,
$$
\|\xi\|_{\psi_2} \frac{\E\|G\|_{H_s}}{\sqrt{N}}  \lesssim s^2,
$$
leading to \eqref{eq:multi-subgaussian-cond}.

In a similar way, the second part of Theorem \ref{thm:subgaussian-quad-and-multi} leads to the choice of $s_Q$.

Combining these observations, the following is a bound on the estimation problem for the squared loss in a subgaussian setup, and which achieves the minimax rates in rather general situations (see \cite{LM13} for more details).

\begin{Theorem} \label{thm:LM-ERM} \cite{LM13}
For every $L \geq 1$ there exist constants $c_1,c_2,c_3$ and $c_4$ that depend only on $L$ for which the following holds. Let $F$ be a convex, centrally symmetric, $L$-subgaussian class of functions and assume that  $\|Y-f^*(X)\|_{\psi_2}\leq \sigma$. Set $\eta_1= c_1/\sigma$ and $\eta_2 = c_2$, and put $s_M=s_M(\eta_1)$ and $s_Q=s_Q(\eta_2)$.
\begin{description}
\item{1.} If $\sigma \geq c_3s_Q$ then with probability at least $1-4\exp(-c_4N\eta_1^2 s_M^2)$,
$ \|\hat{f}-f^*\|_{L_2} \leq s_M^2$.
\item{2.} If $\sigma \leq c_3s_Q$ then with probability at least $1-4\exp(-c_4N\eta_2^2)$, $\|\hat{f}-f^*\|_{L_2} \leq s_Q^2$.
\end{description}
\end{Theorem}
\begin{Remark}
Note that $\eta_1 \sqrt{N} \sim s_M^{-2} \E\|G\|_{F \cap {s_M}D}$, and thus $\eta_1^2 N s_M^2 \sim k_{F \cap {s_M}D}$. Therefore, the probability estimate in (1) is $1-4\exp(-c k_{F \cap {s_M}D})$.
\end{Remark}

To put Theorem \ref{thm:LM-ERM} in some perspective, the two fixed points that govern the error rates are precisely as expected: $s_Q(c_2)$ is an intrinsic parameter because it depends only on the class $F$ and not on the target $Y$. It measures the statistical complexity of the class $F$ -- it is an upper estimate (which is often sharp) on the $L_2$ diameter of the version space\footnote{it should be noted that as a bound on the diameter of the version space, controlling $s_Q$ leads to interesting results in other problems with a geometric flavour, like estimates on the diameter of the kernel of a random matrix (see, for example, \cite{MR845980,Shahar-Gelfand}), approximate reconstruction \cite{MR2373017}, estimates on the smallest singular value of certain random matrices, etc. All these problems share a common thread -- that a certain random operator or a random sampling method is injective and stable. For example, a typical result for a convex set $T \subset \R^n$ is based on the following: if $\Gamma=N^{-1/2}\sum_{i=1}^N \inr{X_i,\cdot}e_i$ is a random matrix with independent rows distributed according to $X$, then $\|\Gamma s - \Gamma t\|_{\ell_2^N} \geq c\|s-t\|_{\ell_2^n}$ provided that $s,t \in T$ and that $\|s-t\|_{\ell_2^n}$ is sufficiently large. This leads to an estimate on the random Gelfand width of $T$, and when $T=B_2^n$ to a lower bound on the smallest singular value of $\Gamma$.} and in particular, controls the estimation error in a noise-free situation.

The multiplier process has a geometric interpretation: for every $(X_i)_{i=1}^N$ and $(\xi_i)_{i=1}^N=(f^*(X_i)-Y_i)_{i=1}^N$ (that need not be independent of the $X_i$'s), it measures the width (or correlation) of the set
$$
\left\{\left(f(X_i)\right)_{i=1}^N : f \in F, \ \|f-f^*\|_{L_2} \leq s\right\}
$$
relative to the weighted Bernoulli random vector $(\eps_i \xi_i)_{i=1}^N$. The width clearly increases with the length of the random vector, and so, with noise level of the problem, captured here by $\|f^*(X)-Y\|_{\psi_2}$. Therefore, once enough noise is introduced to the problem, the impact of the multiplier process increases and $s_M(c_1/\sigma)$ becomes dominant.

\vskip0.5cm

Observe that there is a link between the two parameters and the structure of the excess loss. Not only are there two noise regimes, each captured by a different parameter, but also each regime originates from a different part of the excess loss functional: the intrinsic parameter from the quadratic part and the external parameter from the multiplier component. The transition between a low-noise problem and a high-noise one occurs based on the dominating component of the loss.

\subsection{Towards a general theory - preliminary remarks}
If one wishes, as we do, to extend the results from the subgaussian case outlined above to a more general scenario, one must overcome two main obstacles.

First, one has to modify the concentration-based argument used in Theorem \ref{thm:LM-ERM}, simply because versions of Theorem \ref{thm:subgaussian-quad-and-multi} are false in heavy-tailed situations; second, one must find a way of studying general loss functions, rather than the squared loss.

Bypassing concentration-based arguments is possible thanks to the small-ball condition. \begin{Definition}
A random variable $Z$ satisfies a small-ball condition with constants $\kappa>0$ and $0<\eps<1$ if
$$
Pr(|Z| \geq \kappa \|Z\|_{L_2}) \geq \eps.
$$
A class of functions $F$ defined on the probability space $(\Omega,\mu)$ satisfies a small-ball condition with constants $\kappa$
and $0<\eps<1$ if for every $f \in F$,
$$
Pr(|f| \geq \kappa \|f\|_{L_2}) \geq \eps.
$$
\end{Definition}
This small-ball condition has been introduced in the context of estimation problems in \cite{Shahar-COLT}, and is the most important feature of our presentation. Being a rather weak assumption that is almost universally satisfied (see \cite{Shahar-COLT} for some examples), it serves as a replacement for concentration that comes almost free of charge.

\vskip0.5cm

As for more general loss functions, the need for a theory that can handle those extends beyond the obvious reason -- that the square loss is not the only loss used in applications. A more subtle and interesting reason has to do with the existence of outliers.

The combination of the rapid growth of the squared loss with heavy-tailed sampling inevitably leads to outliers -- sample points that are misleading (because of the heavy tails) and have a significant impact on ERM (because the loss grows quickly).

It is highly desirable to find a way of removing the ill-effects of outliers, and we will show that one possibility is choosing a loss that is calibrated to fit the noise level and the intrinsic structure of the underlying class.

\vskip0.5cm

As noted above, the most extensively studied loss is the squared loss $\ell(t)=t^2$, which is also the most basic example of a strongly convex loss:
\begin{Definition} \label{def:strictly-convex}
A function $\ell$ is strongly convex in the interval $I$ with a constant $c_0>0$ if
$$
\ell(y) \geq \ell(x)+\ell^\prime(x)(y-x)+\frac{c_0}{2}(y-x)^2
$$
for every $x,y \in I$.
\end{Definition}
Clearly, if $ \inf_{x \in \R} \ell^{\prime \prime}(x) \geq c>0$ then $\ell$ is strongly convex in $\R$ with a constant $c$.
\vskip0.5cm

If one wishes the loss to be convex, its growth from any point must be at least linear. Therefore, it seems natural to consider loss functions that are strongly convex in an interval around zero, thus mimicking the local behaviour of the squared loss; and, away from zero, exhibit a linear, or almost linear growth, hopefully limiting the negative effect of outliers.

Typical examples of such losses are the Huber loss with parameter $\gamma$, defined by
\begin{equation} \label{eq:Huber-loss}
\ell_\gamma(t)=
\begin{cases}
\frac{1}{2}t^2 & {\rm if} \ |t| \leq \gamma
\\
\gamma |t|-\frac{\gamma^2}{2} & {\rm if} \ |t| > \gamma,
\end{cases}
\end{equation}
and a version of the logistic loss,
\begin{equation} \label{eq:log-loss}
\ell(t)=-\log\left(\frac{4\exp(t)}{(1+\exp(t))^2}\right),
\end{equation}
which is strongly convex in any bounded interval, but with a constant that decays exponentially to zero with the interval's length (because $\ell^{\prime \prime}(t)= 2\exp(t)/(\exp(t)+1)^2$).

\vskip0.5cm

The general framework that will be developed here aims at going beyond the subgaussian theory and the squared loss:
\begin{description}
\item{$\bullet$} We will extend the natural decomposition of the squared excess loss to more general losses, leading to a better understanding of the important features of the loss, and to the correct notions of `high-noise' and `low-noise' regimes.
\vskip0.3cm
\item{$\bullet$} We will develop suitable one-sided lower bounds that are based on a small-ball argument, replacing the restrictive concentration-based two-sided estimates.
\vskip0.3cm
\item{$\bullet$} We will explain how the choice of the loss may be used to address the outliers issue, with a particularly striking effect when the class is well behaved and the target is heavy tailed.
\end{description}

\subsection{Some notation}
Throughout the article, absolute constants are denoted by $c_1,c_2,...$; their value may change from line to line. We write $A \lesssim B$ if there is an absolute constant $c_1$ for which $A \leq c_1 B$, and $A \sim B$ if $c_1 A \leq B \leq c_2 A$ for absolute constants $c_1$ and $c_2$.  $A \lesssim_r B$ or $A \sim_r B$ means that the constants depend on some parameter $r$. $\kappa_0$, $\kappa_1$,... etc, denote constants whose value remains fixed.

Given a probability measure $\mu$, set $D=B(L_2(\mu))$ to be the unit ball of $L_2(\mu)$, let $rD$ be the ball of radius $r$ and put $rD_f$ to be the ball centred in $f$ and of radius $r$. $S(L_2)$ denotes the unit sphere in $L_2$ and $S^{n-1}$ is the Euclidean sphere in $\R^n$. Throughout the article we will not specify the $L_2$ space on which the functions in question belong, as that will be clear from the context.

For $\alpha \geq 1$, $L_{\psi_\alpha}$ is the Orlicz space of all measurable functions, for which the $\psi_\alpha$ norm, defined by
$$
\|f\|_{\psi_\alpha}=\inf\left\{ c>0 : \E \exp(|f/c|^\alpha) \leq 2 \right\},
$$
is finite. Some basic facts
on Orlicz spaces may be found, for example, in \cite{vanderVaartWellner}.

Given $1 \leq p < \infty$, let $B_p^n = \{x \in \R^n : \sum_{i=1}^n |x_i|^p \leq 1 \}$ be the unit ball in the space $\ell_p^n$, with the obvious modification when $p=\infty$.
\vskip0.4cm
A class of functions $H$ is star-shaped around $0$ if for every $h \in H$ and every $\lambda \in [0,1]$, $\lambda h \in H$. In other words, if $h \in H$ then $H$ contains the entire interval connecting $h$ to $0$.

It is straightforward to verify that if $F$ is convex and $f \in F$ then $H_f=F-f=\{h-f : h \in F\}$ is star-shaped around $0$.

A class that is star-shaped around zero has some regularity. The star-shape property implies that if $r < \rho$, then $H \cap r S(L_2)$ contains a `scaled-down' version of $H \cap \rho S(L_2)$. Indeed, if $h \in H \cap \rho S(L_2)$ and since $r/\rho \in [0,1]$, it follows that $(r/\rho)h \in H \cap r S(L_2)$. In particular, normalized `layers' of a star-shaped class become richer the closer the layer is to zero.

Finally, if $A$ is a finite set, we denote by $|A|$ its cardinality.

\subsection{The Organization of the article}
The rest of the article is arranged as follows. In Section \ref{sec:scheme} we will present the new scheme for dealing with a general loss function. Then, in Section \ref{sec:est-monotone} and Section \ref{sec:emp-small-ball} we will define $r_Q$ -- the intrinsic complexity of the class, and use the small-ball condition to derive uniform lower bounds on the `quadratic component' of a general loss function.

Next, in Section \ref{sec:orcle}, we will identify the external parameter, $r_M$, that captures the interaction of the class, the noise and the loss. This will be followed by proofs of the main results of this article --  a solution of Problem \ref{qu:main} for a general loss, without any tail restrictions on the class, nor on the target, while satisfying the entire `wish-list' outlined earlier.

Finally, in Section \ref{sec:pre-examples} we will show how the main results may be used for three loss functions (the squared loss, the logistic loss and the Huber loss). Moreover, we will show that a wise choice of the loss may be used to treat the issue of outliers in heavy-tailed scenarios.

As concrete examples, we will present bounds on ${\cal E}_p$ and ${\cal E}_e$ when $F=\R^n$ or when $F=rB_1^n$, both viewed as classes of linear functionals on $\R^n$.

As will be explained in Section \ref{sec:pre-examples}, one of the outcomes of the general theory developed here is that (roughly and somewhat inaccurately put) by selecting a loss that grows linearly in the ray $[c\max\{r_Q,\|\xi\|_{L_2}\},\infty)$ and that is strongly convex in the interval $[0,c\max\{\|\xi\|_{L_2},r_Q\})$, one obtains the same error rates as if $\xi$ were a gaussian variable, independent of $X$. In particular, this shows that the impact of outliers generated because of a heavy-tailed target can be negated using a well-calibrated loss that fits both the intrinsic complexity of the class (via $r_Q$) and the level of the noise (via $\|\xi\|_{L_2}$).

\section{The general scheme -- beyond the squared loss} \label{sec:scheme}
As explained earlier, the analysis of ERM is based on exclusion: showing that a large (random) part of the class cannot contain the empirical minimizer because the empirical risk is positive for functions that belong to it.

The way one excludes parts of the class depends on the type of estimate one would like to obtain. For the estimation problem (an estimate on $\|\hat{f}-f^*\|_{L_2}$), it makes sense to `localize' according to the $L_2$ distance from $f^*$. However, when the goal is a prediction problem, where one must show that with high probability, the conditional expectation $\E {\cal L}_{\hat{f}}$ is small, the natural localization of $F$ is according to level sets of the excess risk functional $\E {\cal L}_f$.

In this section we will present a scheme leading to bounds on both prediction and estimation, which is based on the decomposition of the loss to a multiplier component and a quadratic one. This is achieved via a second order Taylor expansion of $\ell$ around the value $\xi=f^*(X)-Y$ when $\ell$ is smooth enough, and with some modifications when it is not.

If $\ell$ has a second derivative then for every $(X,Y)$ there is a mid-point $Z$ for which
\begin{align*}
{\cal L}_f (X,Y) & = \ell(f(X)-Y)-\ell(f^*(X)-Y)
\\
& = \ell^\prime(\xi)(f-f^*)(X)+\frac{1}{2} \ell^{\prime \prime}(Z)(f-f^*)^2(X)=(1)+(2).
\end{align*}
One may exclude $F^\prime \subset F$ by showing that the empirical mean of the quadratic term (2) is positive on $F^\prime$, while the empirical mean of the multiplier component (1) cannot be very negative there.

If $\ell$ does not have a second derivative everywhere, one may modify this decomposition by noting that for every $x_1$ and $x_2$,
$$
\ell(x_2)-\ell(x_1)=\int_{x_1}^{x_2} \ell^\prime(w)dw=\ell^\prime(x_1)(x_2-x_1)+\int_{x_1}^{x_2} (\ell^\prime(w)-\ell^\prime(x_1))dw.
$$
Therefore, when applied to $(X,Y)$ and a fixed $f \in F$, the quadratic component in the decomposition is
$$
\int_\xi^{\xi+(f-f^*)(X)} \left(\ell^\prime(w)-\ell^\prime(\xi)\right)dw.
$$
In particular, it is straightforward to show that if $\ell$ is twice differentiable in $\R$, except, perhaps, in $\pm x_0$, then for every $X,Y$ one has
$$
{\cal L}_f (X,Y) \geq \ell^\prime(\xi)(f-f^*)(X)+\frac{1}{16} \ell^{\prime \prime}(Z)(f-f^*)^2(X)
$$
for a well-chosen midpoint $Z$.

\begin{Definition} \label{M-f-Q-f}
For every $f,f^* \in F$ and $(X,Y)$, set
$$
{\cal M}_{f-f^*}(X,Y) = \ell^\prime(\xi)(f-f^*)(X),
$$
and put
$$
{\cal Q}_{f-f^*}(X,Y)=\int_\xi^{\xi+(f-f^*)(X)} \left(\ell^\prime(w)-\ell^\prime(x)\right)dw,
$$
representing the multiplier component of the excess loss and the quadratic one, respectively.
\end{Definition}

A structural assumption that will be needed throughout this exposition is the following:
\begin{Assumption} \label{ass:modified-Bernstein}
Assume that for every $f \in F$,
$$
\E \ell^\prime(\xi) (f-f^*)(X) \geq 0.
$$
\end{Assumption}
Assumption \ref{ass:modified-Bernstein} is not really restrictive:
\begin{description}
\item{$\bullet$} If $\xi(X,Y)=f^*(X)-Y$ is independent of $X$ (e.g., when $Y=f_0(X)-W$ for some unknown $f_0 \in F$ and an independent, symmetric random variable $W$), then $\E \ell^\prime(\xi) (f-f^*)(X)=0$, because $\ell^\prime$ is odd.
\item{$\bullet$} If $F$ is a convex class of functions and $\ell$ satisfies minimal integrability conditions, then $\E \ell^\prime(\xi) (f-f^*)(X) \geq 0$ for every $f \in F$. Indeed, if there is some $f_1 \in F$ for which $\E\ell^\prime(\xi)(f_1-f^*)(X) < 0$, then by considering $f_\lambda = \lambda f_1 + (1-\lambda)f^* \in F$ for $\lambda$ close to $0$,
$$
\E\ell(f_\lambda(X)-Y) < \E \ell(f^*(X)-Y),
$$
which is impossible.
\end{description}

Under Assumption \ref{ass:modified-Bernstein}, given a sample $(X_i,Y_i)_{i=1}^N$ and $f \in F$, there are mid-points $Z_i$ that fall between $f^*(X_i)-Y_i$ and $f(X_i)-Y_i=(f-f^*)(X_i)+\xi_i$, for which
\begin{align} \label{eq:Taylor-expnation}
P_N {\cal L}_f = & \frac{1}{N} \sum_{i=1}^N \ell(f(X_i)-Y_i) - \ell(f^*(X_i)-Y_i) \nonumber
\\
\geq & \frac{1}{N} \sum_{i=1}^N \ell^\prime(\xi_i) (f-f^*)(X_i) + \frac{1}{2N} \sum_{i=1}^N \ell^{\prime \prime} (Z_i) (f-f^*)^2(X_i) \nonumber
\\
\geq & \left|\frac{1}{N} \sum_{i=1}^N \ell^\prime(\xi_i) (f-f^*)(X_i)-\E \ell^\prime (\xi) (f-f^*)\right| + \E \ell^\prime (\xi) (f-f^*) \nonumber
\\
+ & \frac{1}{2N} \sum_{i=1}^N \ell^{\prime \prime} (Z_i) (f-f^*)^2(X_i).
\end{align}
Assume that on a high-probability event ${\cal A}$, for every $f \in F$,
$$
\left|\frac{1}{N}\sum_{i=1}^N \ell^\prime(\xi_i)(f-f^*)(X_i) - \E \ell^\prime(\xi)(f-f^*)(X)\right| \leq \frac{\theta}{4}\max \left\{\|f-f^*\|_{L_2}^2,r_M^2\right\},
$$
for well chosen values $r_M$ and $\theta$. Assume further that on a high probability event ${\cal B}$, for every $f \in F$ with $\|f-f^*\|_{L_2} \geq r_Q$,
$$
\frac{1}{N}\sum_{i=1}^N \ell^{\prime \prime}(Z_i)(f-f^*)^2(X_i) \geq \theta \|f-f^*\|_{L_2}^2.
$$

\begin{Theorem} \label{thm:estimation-intro}
If $F$ satisfies Assumption \ref{ass:modified-Bernstein}, then on the event ${\cal A} \cap {\cal B}$, $\|\hat{f}-f^*\|_{L_2} \leq \max\{r_M,r_Q\}$.
\end{Theorem}
\proof By Assumption \ref{ass:modified-Bernstein} and \eqref{eq:Taylor-expnation},
\begin{align*}
P_N{\cal L}_f \geq & \frac{1}{2N}\sum_{i=1}^N \ell^{\prime \prime}(Z_i)(f-f^*)^2(X_i)
\\
- & \left|\frac{1}{N}\sum_{i=1}^N \ell^\prime(\xi_i)(f-f^*)(X_i) - \E \ell^\prime(\xi)(f-f^*)(X)\right|.
\end{align*}
Hence, on the event ${\cal A} \cap {\cal B}$, if $\|f-f^*\|_{L_2} \geq \max\{r_M,r_Q\}$ then $P_N{\cal L}_f \geq (\theta/4)\|f-f^*\|_{L_2}^2 >0$, and $f$ cannot be an empirical minimizer.
\endproof
Therefore, to resolve the estimation problem it suffices to identify $r_M$ and $r_Q$ for which the event ${\cal A} \cap {\cal B}$ is sufficiently large.

\vskip0.5cm

Turning to the prediction problem, there is an additional assumption that is needed, namely, that $\E{\cal Q}_{f-f^*}$ does not increase too quickly when $f$ is close to $f^*$.

\begin{Assumption} \label{ass:quadratic-for-oracle}
Assume that there is a constant $\beta$ for which, for every $f \in F$ with $\|f-f^*\|_{L_2} \leq \max\{r_M,r_Q\}$, one has
$$
\E {\cal Q}_{f-f^*} \leq \beta \|f-f^*\|_{L_2}^2.
$$
\end{Assumption}

Clearly, if $\ell^\prime$ is a Lipschitz function, one may take $\beta=\|\ell^\prime\|_{{\rm lip}}$; hence, if
$\ell^{\prime \prime}$ exists everywhere and is a bounded function, $\beta \leq \|\ell^{\prime \prime}\|_{L_\infty}$.
Moreover, even when $\ell^{\prime \prime}$ is not bounded, such a $\beta$ exists if the functions $f-f^*$ have well behaved tails relative to the growth of $\ell^{\prime \prime}$. Since the analysis required in these cases is rather obvious, we will not explore this issue further.

\begin{Theorem} \label{thm:prediction-intro}
Assume that $\ell$ satisfies Assumption \ref{ass:modified-Bernstein} and Assumption \ref{ass:quadratic-for-oracle}. Using the notation introduced above, on the event ${\cal A} \cap {\cal B}$ one has
$$
\E {\cal L}_{\hat{f}} \leq 2(\theta+\beta) \max\{r_M^2,r_Q^2\}.
$$
\end{Theorem}
\proof Fix a sample in ${\cal A} \cap {\cal B}$. By Theorem \ref{thm:estimation-intro}, $\|\hat{f}-f^*\|_{L_2} \leq \max\{r_M,r_Q\}$. Thus, it suffices to show that if $\|f-f^*\|_{L_2} \leq \max\{r_M,r_Q\}$ and $\E {\cal L}_f \geq 2(\theta+\beta) \max\{r_M^2,r_Q^2\}$, then $P_N{\cal L}_f >0$; in particular, such a function cannot be an empirical minimizer.

Note that for every $f \in F$, ${\cal L}_f ={\cal M}_{f-f^*} + {\cal Q}_{f-f^*}$ and thus either $\E {\cal L}_f \leq 2\E{\cal M}_{f-f^*}=\E \ell^\prime(\xi)(f-f^*)(X)$ or $\E {\cal L}_f \leq 2\E {\cal Q}_{f-f^*}$.

However, if $f$ satisfies the above, only the first case is possible; indeed, if $\E {\cal Q}_{f-f^*}$ is dominant, then by Assumption \ref{ass:quadratic-for-oracle},
$$
\E {\cal L}_f \leq 2\E {\cal Q}_{f-f^*} \leq 2\beta \|f-f^*\|_{L_2}^2 \leq 2\beta\max\{r_M^2,r_Q^2\},
$$
which is impossible. Therefore, it suffices to treat the case in which $\E {\cal L}_f \leq 2\E \ell^\prime(\xi)(f-f^*)(X)$.

Fix such an $f \in F$. Since $\ell$ is convex, $P_N {\cal L}_f \geq \frac{1}{N}\sum_{i=1}^N \xi_i(f-f^*)(X_i)$, and on ${\cal A} \cap {\cal B}$,
\begin{align*}
P_N {\cal L}_f \geq & \E \ell^\prime(\xi)(f-f^*)(X) - \left|\frac{1}{N} \sum_{i=1}^N \ell^\prime(\xi)(f-f^*)(X_i) - \E \ell^\prime(\xi)(f-f^*)(X)\right|
\\
\geq & \frac{1}{2}\E {\cal L}_f - \frac{\theta}{4} \max\{r_M^2,\|f-f^*\|_{L_2}^2\}
\\
\geq & (\theta+\beta)  \max\{r_M^2,r_Q^2\} - \frac{\theta}{4}\max\{r_M^2,r_Q^2\} >0.
\end{align*}
\endproof

In the following sections we will develop the necessary machinery leading to a uniform lower estimate on the quadratic term $f \to P_N {\cal Q}_{f-f^*}$ and to an upper estimate on the multiplier term $f \to P_N{\cal M}_{f-f^*}$. Combining the two, we will identify the values $r_Q$ and $r_M$, as well as the right choice of $\theta$.

\section{Preliminary estimates} \label{sec:est-monotone}
Let $(Z_i)_{i=1}^N$ be independent copies of a random variable $Z$ and set $(Z_i^*)_{i=1}^N$ to be a monotone non-increasing rearrangement of $(|Z_i|)_{i=1}^N$.

This section is devoted to the derivation of upper and lower estimates on various function of $(Z_i^*)_{i=1}^N$. All the bounds presented here are well-known and straightforward applications of either a concentration inequality for $\{0,1\}$-valued random variables ({\it selectors}) with mean $\delta$, or, alternatively, a rather crude binomial estimate.

Given a property ${\cal P}$ let $\delta_i = \IND_{\{Z_i \in {\cal P}\}}$, the characteristic function of the event that $Z_i$ satisfies property ${\cal P}$. Set $\delta = Pr(Z \in {\cal P})$ and note that $|\{i : Z_i \in {\cal P}\}| = \sum_{i=1}^N \delta_i$. By Bernstein's inequality (see, for example, \cite{vanderVaartWellner,BouLugMass13}),
$$
Pr\left(\left|\frac{1}{N}\sum_{i=1}^N \delta_i  - \delta \right| \leq t \right) \geq 1-2\exp(-cN \min\{t^2/\delta,t\})
$$
for a suitable absolute constant $c$. Hence, taking $t = u \delta$,
\begin{equation} \label{eq:card-Bernstein}
N\delta (1-u) \leq |\{i : Z_i \in {\cal P}\}| \leq N\delta(1+u)
\end{equation}
with probability at least $1-2\exp(-cN\delta\min\{u^2,u\})$.

\vskip0.5cm

The binomial estimate is based on the fact that
$$
Pr(|\{i : Z_i \in {\cal P}\}| \geq k) \leq \binom{N}{k}Pr^k(Z \in {\cal P}) \leq \left(\frac{eN}{k} \cdot Pr(Z \in {\cal P})\right)^k.
$$

\subsection{Tail-based upper estimates} \label{sec:est-upper}
Assume that one has information on $\|Z\|_{L_q}$ for some $q \geq 2$ and set $L = \|Z\|_{L_q}/\|Z\|_{L_2}$. Applying Chebyshev's inequality,
$$
Pr(|Z| \geq w \|Z\|_{L_2}) \leq \frac{\E|Z|^q}{\|Z\|_{L_2}^q w^q} = \frac{L_q^q}{w^q}.
$$
Hence, if ${\cal P}=(|Z| < w\|Z\|_{L_2})$ it follows that one may take $\delta=1-(L/w)^q$, which can be made arbitrarily close to $1$ by selecting $w$ that is large enough. This implies that with high probability, an arbitrary large proportion of $\{|Z_1|,...,|Z_N|\}$ are not very large.

\begin{Lemma} \label{lemma-monotone-before-0}
There exists absolute constants $c_1$ and $c_2$ for which the following holds. Let $Z \in L_2$. For every $0<\eps<1$, with probability at least $1-2\exp(-c_1\eps N)$ there exists a subset $I \subset \{1,...,N\}$, $|I| \geq (1-\eps)N$, and for every $i \in I$,
$$
|Z_i| \leq c_2\eps^{-1/2}\|Z\|_{L_2}.
$$
\end{Lemma}

\proof Fix $\eps$ as above and note that $Pr(|Z| \geq 2\|Z\|_{L_2}/\sqrt{\eps}) \leq \eps/4$. Hence, by a binomial estimate,
\begin{align*}
& Pr(|\{i: |Z_i| \geq 2\|Z\|_{L_2}/\sqrt{\eps}\}| \geq N\eps) \leq \binom{N}{\eps N} Pr^{\eps N}(|Z| \geq 2\|Z\|_{L_2}/\sqrt{\eps})
\\
\leq & \left(\frac{e}{\eps}\right)^{N \eps} \cdot \left(\frac{\eps}{4}\right)^{N\eps} \leq \exp(-c N \eps),
\end{align*}
for a suitable absolute constant $c$.
\endproof

Given a vector $a=(a_i)_{i=1}^N$, the $L_q$ norm of $a$, when considered as a function on the space $\{1,...,N\}$ endowed with the uniform probability measure, is
$$
\|a\|_{L_q^N}=\left(\frac{1}{N}\sum_{i=1}^N |a_i|^q\right)^{1/q}.
$$
 The weak-$L_q$ norm of the vector $a$ is
$$
\|a\|_{L_{q,\infty}^N}=\inf\{c>0: d_a(t) \leq (c/t)^q \ {\rm for \ every \ } t>0\},
$$
where $d_a(t)=N^{-1}|\{i: |a_i|>t\}|$.

The next observation is that sampling preserves the $L_q$ structure of $Z$, in the sense that if $Z \in L_q$, then with high probability, $\|(Z_i)_{i=1}^N\|_{L_{q,\infty}^N} \lesssim \|Z\|_{L_q}$.
\begin{Lemma} \label{lemma-monotone-0}
Let $1 \leq q \leq r$. If $Z \in L_r$, $u \geq 2$ and $1 \leq k \leq N/2$, then
$$
Z_k^* \leq u(N/k)^{1/q}\|Z\|_{L_r}
$$
with probability at least $1-u^{-kr}\left(\frac{eN}{k}\right)^{-k((r/q)-1)}$.

In particular, with probability at least $1-2u^{-r}N^{-((r/q)-1)}$,
$$
\|(Z_i)\|_{L_{q,\infty}^N} \leq u\|Z\|_{L_r}.
$$
\end{Lemma}
\proof
Let $\eta=(r/q)-1$, fix $1 \leq k \leq N/2$ and set $v>0$ to be named later. The binomial estimate implies that
\begin{align*}
& Pr(Z_k^* \geq v(eN/k)^{(1+\eta)/r} \|Z\|_{L_r}) \leq \binom{N}{k} Pr^k (|Z| \geq v(eN/k)^{(1+\eta)/r}\|Z\|_{L_r})
\\
\leq & \left(\frac{eN}{k}\right)^k \left(\frac{k}{eN}\right)^{(1+\eta)k} \cdot v^{-kr} =\left(\frac{e N}{k}\right)^{-\eta k} v^{-k r}.
\end{align*}
Hence, for $v=u (eN/k)^{1/q - (1+\eta)/r}$,
$$
Z_k^* \leq u(N/k)^{1/q} \|Z\|_{L_r}
$$
with probability at least
$$
1-u^{-kr}\left(\frac{eN}{k}\right)^{k((r/q)-1)}.
$$
The second part of the claim follows by summing up the probabilities for $k \leq N/2$, using that $Z_k^* \leq Z_{N/2}^*$ for $k \geq N/2$ and that $(u^{-kr})_{k=1}^{N/2}$ is a geometric progression.

\begin{Remark}
The upper estimates presented above are based on the fact that if $Z \in L_q$ and $p \leq q$ then $Pr(|Z| \geq t\|Z\|_{L_p})$ can be made arbitrarily close to $1$ for a choice of $t$ that is independent of $Z$. Similar arguments are true if one simply assumes that $Pr(|Z| \geq t)<\eps$, even without moment assumptions. Of course, under such an assumption one has no information whatsoever on the largest $\eps N$ coordinates of $(|Z_1|,....,|Z_N|)$, but rather, only on a certain proportion that is slightly smaller than $(1-\eps)N$ of the coordinates.

Also, observe that $\|(Z_i^*)_{i \geq j}\|_{L_{q,\infty}^N} \lesssim \|Z\|_{L_q}$ with a probability estimate that improves exponentially in $j$.
\end{Remark}

\subsection{Lower estimates using a small-ball property} \label{sec:lower-est}
A similar line of reasoning to the one used above is true for lower estimates. Because the applications considered below require many of the $|Z_i|$'s to be at least of the order of $\|Z\|_{L_2}$, that norm is used as a point of reference in the definition of the small-ball condition, that
$$
Pr(|Z| \geq \kappa \|Z\|_{L_2}) \geq \eps
$$
for constants $\kappa$ and $0<\eps<1$.

Of course, the notion of `small-ball' can be modified to fit other norms, as well as situations in which $Z$ does not have any moments.

\vskip0.5cm

The small-ball condition is a weak assumption and we refer the reader to \cite{Shahar-COLT} for several examples of classes of functions for which it holds. One generic situation in which the small-ball condition is satisfied is when $Z$ is a random variable for which $\|Z\|_{L_q} \leq L \|Z\|_{L_2}$ for some $q>2$. It follows from the Paley-Zygmund inequality (see, e.g. \cite{MR1666908}) that $Z$ satisfies a small-ball condition with constants $\kappa$ and $\eps$ that depend only on $L$ and $q$. Moreover, if $Z$ is also mean-zero and $W=(Z_1,...,.Z_n)$ is a vector consisting of independent copies of $Z$, then for every $x \in \R^n$, $\inr{x,W}=\sum_{i=1}^N x_i Z_i$ satisfies the small-ball condition with constants that depend only on $L$ and $q$ (and not on $x$ or on the dimension $n$). Therefore, if $\mu_W$ is the measure endowed on $\R^n$ by the random vector $W$, any class of linear functionals on the measure space $(\R^n,\mu_W)$ satisfies the small-ball condition with constants that depend only on $L$ and $q$.

\begin{Lemma} \label{lemma:empirical-small-ball}
There exists an absolute constant $c$ for which the following holds.
Assume that $Z$ satisfies a small-ball condition with constants $\kappa_0$ and $0<\eps<1$ and let $(Z_i)_{i=1}^N$ be independent copies of $Z$. Then, with probability at least $1-2\exp(-cN\eps)$, there is a subset $I$ of $\{1,...,N\}$ of cardinality at least $(3/4)\eps N$, and for every $i \in I$, $|Z_i| \geq \kappa_0 \|Z\|_{L_2}$.
\end{Lemma}

The proof, which we omit, is an immediate application of Bernstein's inequality for the i.i.d. selectors $\delta_i = \IND_{\{|Z_i| \geq \kappa \|Z\|_{L_2}\}}$, and for the choice of $u=1/4$ in \eqref{eq:card-Bernstein}. Naturally, at a price of a weaker probability estimate, the constant $3/4$ can be made arbitrarily close to $1$.

Combining the upper estimate from Lemma \ref{lemma-monotone-before-0} and lower one from Lemma \ref{lemma:empirical-small-ball} yields the following corollary:
\begin{Corollary} \label{cor:emp-upper-and-lower}
There exist absolute constants $c_1$ and $c_2$ for which the following holds.
Assume that $Z \in L_2$ and that it satisfies the small-ball condition for constants $\kappa_0$ and $\eps$. Then, with probability at least $1-2\exp(-c_1\eps N)$,
there is $J \subset \{1,...,N\}$, $|J| \geq  \eps N/2$ and for every $j \in J$,
$$
\kappa_0 \|Z\|_{L_2} \leq |Z_j| \leq c_2 \|Z\|_{L_2}/\sqrt{\eps}.
$$
\end{Corollary}
Corollary \ref{cor:emp-upper-and-lower} allows one to control the behaviour of $(Z_i)_{i=1}^N$ on a subset of $\{1,...,N\}$ of cardinality $\sim \eps N$, and with exponentially high probability. Moreover, by modifying $c_1$ and $c_2$, the cardinality of $J$ can be made arbitrarily close to $\eps N$.
\begin{Remark}
Note that by the union bound, a version of Corollary \ref{cor:emp-upper-and-lower} holds uniformly for a collection of $\exp(c_1N\eps/2)$ random variables with probability at least $1-2\exp(-c_1N\eps/2)$ -- an observation that will be used extensively in what follows.
\end{Remark}

\section{A uniform estimate on the quadratic process} \label{sec:emp-small-ball}
The goal of this section is to study the structure of a typical coordinate projection of a class $H$, $P_\sigma H = \{ (h(X_i))_{i=1}^N : h \in H\}$, and show that with high probability, for every function in $H$ of sufficiently large $L_2$ norm, most of the coordinates of $P_\sigma h$ are of the order of $\|h\|_{L_2}$. Such a result is an extension of the `lower part' of Corollary \ref{cor:emp-upper-and-lower} from a single function to a class of functions that is not very big in some sense. The class we will focus on later is $H_{f^*}=\{f-f^* : f \in F\}$.

Given $H \subset L_2(\mu)$ let $\{G_h : h \in H\}$ be the canonical gaussian process indexed by $H$ with a covariance structure endowed by $L_2(\mu)$. Recall that if $H^\prime \subset H$, $\E\|G\|_{H^\prime} = \E \sup_{h \in H^\prime} G_h$.

\begin{Definition} \label{def:r-Q}
Given a class of functions $H \subset L_2(\mu)$, a sample size $N$ and positive constants $\zeta_1$ and $\zeta_2$ set
\begin{equation} \label{eq:gauss-fixed-point}
r_{1,Q}(H,N,\zeta_1)=\inf\left\{ r>0 : \E \|G\|_{H \cap rD} \leq \zeta_1 r \sqrt{N} \right\},
\end{equation}
and put
\begin{equation} \label{eq:emp-fixed-point}
r_{2,Q}(H,N,\zeta_2)=\inf\left\{ r>0 : \E\sup_{H \cap r D} \left|\frac{1}{\sqrt{N}} \sum_{i=1}^N \eps_i h(X_i) \right| \leq \zeta_2 r \sqrt{N}\right\},
\end{equation}
where $D$ is, as always, the unit ball of $L_2(\mu)$.

When the class $H$ and sample size $N$ are obvious from the context, we will denote the fixed points by $r_{1,Q}(\zeta_1)$ and $r_{2,Q}(\zeta_2)$ respectively.

Finally, set
$$
r_Q(\zeta_1,\zeta_2)=r_Q(H,N,\zeta_1,\zeta_2) = \max\{r_{1,Q}(\zeta_1),r_{2,Q}(\zeta_2)\}.
$$
\end{Definition}

By a straightforward application of the Central Limit Theorem, one may show that if $H$ consists of mean-zero functions then
$$
\E \|G\|_{H} \lesssim \limsup_{N \to \infty} \E \sup_{h \in H} \left|\frac{1}{\sqrt{N}} \sum_{i=1}^N \eps_i h(X_i) \right|.
$$
therefore, $r_{2,Q}$ is larger than $r_{1,Q}$, at least asymptotically.

For a fixed $N$, comparing the two parameters is more difficult. In one direction, one has the following lower bound:
\begin{Lemma} \label{lemma:small-ball-implies-gaussian} \cite{shahar-almo-iso-small-ball}
Let $H \subset L_2$ be a class of functions and assume that for every $h_1,h_2 \in H$, $Pr(|h_1-h_2| \geq \kappa \|h_1-h_2\|_{L_2}) \geq \eps$. Then
$$
\E \sup_{h \in H} \left|\frac{1}{\sqrt{N}}\sum_{i=1}^N \eps_i h(X_i) \right| \geq c_1(\eps) \kappa \sup_{H_m} \E\|G_h\|_{H_m}
$$
where the supremum is taken with respect to all subsets of $H$ of cardinality $m \leq \exp(c_2(\eps)N)$.
\end{Lemma}
Note that when the gaussian process $\{G_h : h \in H\}$ is continuous, $N$ is sufficiently large and $H_m$ is selected to be a maximal separated subset of $H$ of cardinality $m$, then $\E \|G\|_{H_m} \geq (1/2) \E\|G\|_H$. Hence, in that range, the expectation of the Bernoulli process indexed by a coordinate projection of $H$ dominates $\E\|G\|_H$.

On the other hand, a standard chaining argument combined with the Majorizing Measures Theorem shows that if $H$ is an $L$-subgaussian class then
$$
\E\sup_{h \in H} \left|\frac{1}{\sqrt{N}} \sum_{i=1}^N \eps_i h(X_i) \right| \lesssim L \E\|G\|_{H}
$$
(see, e.g. \cite{shahar-psi2} and the manuscript \cite{MR3184689} as a general reference for chaining methods).

Thus, the two complexity terms are not that far apart when $H$ is an $L$-subgaussian class.

\vskip0.5cm

If $H$ is star-shaped around $0$, it is straightforward to show that when $r > r_{1,Q}(\zeta_1)$, one has
$$
\E \|G\|_{H \cap r D} \leq \zeta_1 r \sqrt{N},
$$
while if $r<r_{1,Q}(\zeta_1)$,
$$
\E \|G\|_{H \cap r D} \geq \zeta_1 r \sqrt{N}.
$$
A similar observation is true for $r_{Q,2}(\zeta_2)$.

\vskip0.5cm

The following is the main technical tool needed for the study of the quadratic component.
\begin{Theorem} \label{thm:small-ball-basic}
There exist absolute constants $c_0,c_1,c_2,c_3,c_4$ and $c_5$ for which the following holds. Let $H$ be a class of functions that is star-shaped around $0$ and that satisfies a small-ball condition with constants $\kappa_0$ and $\eps$. If $\zeta_1=c_1\kappa_0\eps^{3/2}$, $\zeta_2=c_2 \kappa_0\eps$ and  $r>r_Q(\zeta_1,\zeta_2)$,  there is $V_r \subset H \cap rS(L_2)$ and an event $\Omega^\prime$ of probability at least $1-2\exp(-c_0\eps^2 N)$, with the following properties:
\begin{description}
\item{1.} $|V_r| \leq \exp(c_3 \eps N)$ for $c_3 \leq 1/1000$.
\item{2.} On the event $\Omega^\prime$, for every $v \in V_r$ there is a subset $I_v \subset \{1,...,N\}$, $|I_v| \geq \eps N/2$ and for every $i \in I_v$,
    $$
     \kappa_0 r \leq |v(X_i)| \leq c_4r/\sqrt{\eps}.
    $$
\item{3.} On the event $\Omega^\prime$, for every $h \in H \cap r S(L_2)$ there is some $v \in V_r$ and a subset $J_h \subset I_v$, consisting of at least $3/4$ of the coordinates of $I_v$ (and in particular, $|J_h| \geq \eps N/4$), and for every $j \in J_h$,
    $$
    (\kappa_0/2)\|h\|_{L_2} \leq |h(X_j)| \leq c_5 (\kappa_0+1/\sqrt{\eps})\|h\|_{L_2}
    $$
and
$$
{\rm sgn}(h(X_j))={\rm sgn}(v(X_j)).
$$
\end{description}
\end{Theorem}
The idea of the proof is to find an appropriate net in $H \cap r S(L_2)$ (the set $V_r$), and show that each point in the net has many `well-behaved' coordinates in the sense of (2). Also, if $\pi h$ denotes the best approximation of $h \in H \cap r S(L_2)$ in $V_r$ with respect to the $L_2$ norm, then
$$
\sup_{h  \in H \cap r S(L_2)} \frac{1}{N} \sum_{i=1}^N |h-\pi h|(X_i)
$$
is not very big, showing that $|(h-\pi h)(X_i)|$ cannot have too many large coordinates. Since $h(X_i) = (\pi h)(X_i) + (h-\pi h)(X_i)$, the first term is dominant on a proportional number of coordinates, leading to (3).

\proof Recall that by Corollary \ref{cor:emp-upper-and-lower}, if $Z \in L_2$ satisfies the small-ball condition with constants $\kappa_0$ and $\eps$ then with probability at least $1-2\exp(-c_1\eps N)$,
there is $I \subset \{1,...,N\}$, $|I| \geq  \eps N/2$ and for every $i \in I$,
$$
\kappa_0 \|Z\|_{L_2} \leq |Z_i| \leq c_2 \|Z\|_{L_2}/\sqrt{\eps}.
$$
Fix $\zeta_1$ and $\zeta_2$ to be named later, let $r>r_Q(\zeta_1,\zeta_2)$ and set $V_r \subset H \cap r S(L_2)$ to be a maximal $\eta$-separated set whose cardinality is at most $\exp(c_1^\prime \eps N/2)$, for $c_1^\prime=\min\{c_1,1/500\}$. Therefore, by Corollary \ref{cor:emp-upper-and-lower} and the union bound, it follows that with probability at least $1-2\exp(-c_1 \eps N/2)$ for every $v \in V_r$ there is a subset $I_v$ as above, i.e., $|I_v| \geq \eps N/2$ and for every $i \in I_v$,
$$
\kappa_0 r = \kappa_0 \|v\|_{L_2} \leq |v(X_i)| \leq c_2 \|v\|_{L_2}/\sqrt{\eps} = c_2 r/\sqrt{\eps}.
$$
By Sudakov's inequality (see, e.g. \cite{MR1036275,LT:91,MR1720712}) and since $r \geq r_{Q,1}(\zeta_1)$,
$$
\eta \leq c_3 \frac{\E \|G\|_{H \cap r S(L_2)}}{\sqrt{c_1^\prime N \eps /2}} \leq (c_4 \zeta_1/\sqrt{\eps}) r
$$
for $c_4=\sqrt{2}c_3/\sqrt{c_1^\prime}$.

For every $h \in H \cap r S(L_2)$, let $\pi h \in V_r$ for which $\|h-\pi h\|_{L_2} \leq \eta$, set $u_h=\IND_{\{|h-\pi h| > \kappa_0r/2\}}$ and put
$$
U_r=\{ u_h : h \in H \cap r S(L_2) \}.
$$
Let $\phi(t)=t/(\kappa_0 r/2)$ and note that pointwise, for every $u_h \in U_r$, $u_h(X) \leq \phi(|h-\pi h|(X))$.

Applying the Gin\'{e}-Zinn symmetrization theorem and recalling that $r >r_{Q,2}(\zeta_2)$, one has
\begin{align*}
& \E \sup_{u_h \in U_r} \frac{1}{N} \sum_{i=1}^N u_h(X_i) \leq \E \sup_{h \in H \cap r S(L_2)} \frac{1}{N}\sum_{i=1}^N \phi(|h-\pi h|(X_i))
\\
\leq & \E \sup_{h \in H \cap rS(L_2)} \left|\frac{1}{N}\sum_{i=1}^N \phi(|h-\pi h|(X_i)) - \E \phi(|h-\pi h|(X_i)) \right| + \sup_{h \in H \cap r S(L_2)} \E \phi(|h-\pi h|)
\\
\leq & \frac{4}{\kappa_0 r} \cdot \left( \E \sup_{h \in H \cap rS(L_2)} \left|\frac{1}{N} \sum_{i=1}^N \eps_i (h-\pi h)(X_i) \right|+\sup_{h \in H \cap r S(L_2)} \|h-\pi h\|_{L_2} \right)
\\
\leq & \frac{4}{\kappa_0 r} \cdot \left( 2\zeta_2 r + \eta \right) \leq \frac{\eps}{32},
\end{align*}
provided that $\zeta_1 \sim \kappa_0 \eps^{3/2} $ and $\zeta_2 \sim \kappa_0 \eps $.

Let $\psi(X_1,...,X_N)=\sup_{u \in U_r} \frac{1}{N} \sum_{i=1}^N u(X_i)$. By the bounded differences inequality (see, for example, \cite{BouLugMass13}), with probability at least $1-\exp(-c_5t^2)$,
$$
\psi(X_1,...,X_N) \leq \E \psi + \frac{t}{\sqrt{N}}.
$$
Thus, for $t=\eps\sqrt{N}/32$, with probability at least $1-\exp(-c_6 \eps^2 N)$, $\psi(X_1,...,X_N) \leq \eps/16$, implying that for every $h \in H \cap r S(L_2)$,
$$
|\{i : |h-\pi h|(X_i) \leq (\kappa_0/2)r\}| \geq \left(1-\frac{\eps}{16}\right)N.
$$
Recall that $\pi h \in V_r$ and that $|I_{\pi h}| \geq \eps N/2$. Let
$$
J_h=\{j: |h-\pi h|(X_j) \leq (\kappa_0/2)r\} \cap I_{\pi h}
$$
and thus $|J_h| \geq \eps N/4$. Moreover, for every $j \in J_h$,
\begin{equation} \label{eq:signs}
|h(X_j)| \geq |\pi h (X_j)| - |(h-\pi h)(X_j)| \geq \kappa_0 r - (\kappa_0/2)r =(\kappa_0/2)r,
\end{equation}
which also shows that ${\rm sgn}(h(X_j))={\rm sgn}(\pi h(X_j))$.

The upper estimate follows from a similar argument, using that $|h(X_j)| \leq |\pi h (X_j)| + |(h-\pi h)(X_j)|$.
\endproof

\begin{Remark} \label{rem:star-shaped}
Observe that by the star-shape property of $H$, if  $\rho_1> \rho_2$, then
$$
\frac{1}{\rho_1}\left(H \cap \rho_1 S(L_2)\right) \subset \frac{1}{\rho_2}\left(H \cap \rho_2 S(L_2)\right).
$$
Therefore, certain features of $H \cap \rho_2 S(L_2)$ are automatically transferred to $H \cap \rho_1 S(L_2)$, and in particular,  a version of Theorem \ref{thm:small-ball-basic} holds uniformly for every level that is `larger' than $2r_Q(\zeta_1,\zeta_2)$. Indeed, assume that one has chosen $\rho_2 =2r_Q$ in Theorem \ref{thm:small-ball-basic} and fix $h \in H \cap \rho_1 S(L_2)$. By applying Theorem \ref{thm:small-ball-basic} to $h^\prime=(\rho_2/\rho_1)h \in H \cap \rho_2 S(L_2)$ it follows that on the event $\Omega^\prime$ there is a subset $J$ of $\{1,...,N\}$ of cardinality at least $\eps N/4$ on which
$$
|h(X_j)| \geq (\kappa_0/2)\rho_1  \ \ {\rm and} \ \ {\rm sgn}(h(X_j))={\rm sgn}(\pi h^\prime)(X_j).
$$
\end{Remark}

\vskip0.5cm

Next, let $F \subset L_2$ be a convex set, fix $f^* \in F$ and put $H_{f^*}=\{f-f^* : f \in F\}$. Since $H_{f^*}$ is clearly star-shaped around $0$ and $H_{f^*} \subset F-F$ one has:
\begin{Corollary} \label{thm:small-ball-convex}
If $F$ is a convex class of functions, $F-F$ satisfies the small-ball condition with constants $\kappa_0$ and $\eps$, and $r=2r_Q(F-F,N,\zeta_1,\zeta_2)$, then with probability at least $1-2\exp(-c_0\eps^2 N)$, the following holds. For every $f_1,f_2 \in F$ that satisfy  $\|f_1-f_2\|_{L_2} \geq r$, there is a subset $J_{f_1,f_2} \subset \{1,...,N\}$ of cardinality at least $\eps N/4$ and for every $j \in J_{f_1,f_2}$,
    $$
    |(f_1-f_2)(X_j)| \geq (\kappa_0/2)\|f_1-f_2\|_{L_2}.
    $$
In particular, on the same event,
$$
\inf_{\{f \in F : \|f-f^*\|_{L_2} \geq 2r_Q\}} \frac{1}{N}\sum_{i=1}^N \left(\frac{{f-f^*}}{\|f-f^*\|_{L_2}}\right)^2(X_i) \geq \frac{\eps \kappa_0^2}{16}.
$$
\end{Corollary}

\subsection{The quadratic component of the loss} \label{sec:quad-component-loss}
Following the exclusion idea, the aim is to show that the quadratic component of the loss is sufficiently positive. And, although the results will be formulated in full generality, there are three examples that one should keep in mind: First, when $\ell$ is strongly convex; second, a general loss function, assuming that $Y=f_0(X)+W$ for some $f_0 \in F$ and a symmetric random variable $W$ that is independent of $X$; and finally, a situation that is, in some sense, a mixture of the two: a loss function that is guaranteed to be strongly convex only in a neighbourhood of $0$, and without assuming that the noise is independent of $X$.

Throughout this section we will assume that $F$ is a convex class of functions and that $F-F=\{f-h : f,h \in F\}$ satisfies the small-ball condition with constants $\kappa_0$ and $\eps$. Also,
set $r_Q=r_Q(F-F,N,\zeta_1,\zeta_2)$ with the choice of $\zeta_1$ and $\zeta_2$ as in Theorem \ref{thm:small-ball-basic}; namely, $\zeta_1 \sim \kappa_0\eps^{3/2}$ and $\zeta_2 \sim \kappa_0\eps$. Finally, assuming that $\ell^{\prime \prime}$ exists everywhere except perhaps in $\pm x_0$, set for every $0<t_1<t_2$
\begin{equation} \label{eq:def-rho}
\rho(t_1,t_2)=\inf \{\ell^{\prime \prime}(x) : x \in [t_1,t_2], \ x \not = \pm x_0\}.
\end{equation}

\vskip0.5cm

The following lower bound on the quadratic component in the strongly convex case is an immediate application of Corollary \ref{thm:small-ball-convex} and the fact that ${\cal Q}_{f-f^*}(X,Y) \gtrsim \ell^{\prime \prime}(Z)(f-f^*)^2(X)$ for an appropriate mid-point $Z$. Its proof is omitted.
\begin{Theorem} \label{thm:quad-strictly-convex}
There exists an absolute constant $c_1$ for which the following holds.
If $\inf_{x \in \R \backslash \{\pm x_0\}} \ell^{\prime \prime}(x) \geq 2c_0$, then with probability at least $1-2\exp(-c_1 N \eps^2)$, for every $f \in F$ with $\|f-f^*\|_{L_2} \geq 2r_Q$,
$$
P_N {\cal Q}_{f-f^*} \geq \frac{c_0\eps\kappa_0^2}{16} \|f-f^*\|_{L_2}^2.
$$
\end{Theorem}
Theorem \ref{thm:quad-strictly-convex} generalizes a similar result from \cite{Shahar-COLT} for the squared loss.

\vskip0.5cm

Turning to the more difficult problem of a loss that need not be strongly convex, we begin with the case of independent noise.

\begin{Assumption} \label{ass:independent-quadratic}
Assume that $Y=f_0(X)+W$, for a fixed but unknown $f_0 \in F$ and a symmetric random variable $W \in L_2$ that is independent of $X$ and for which
\begin{equation} \label{eq:small-ball-noise}
Pr(|W| \leq \kappa_1\|W\|_{L_2}) \leq \eps/1000.
\end{equation}
\end{Assumption}
Clearly, \eqref{eq:small-ball-noise} is a rather minimal assumption, as a small-ball condition for a single function and at one level holds when the function is absolutely continuous, by selecting the right value $\kappa_1$.

Observe that $f^*=f_0$. Given a sample $(X_i,Y_i)_{i=1}^N$, let $W_i=Y_i-f^*(X_i)$ and set $Z_i$ to be the mid-points in the lower bound on the quadratic component of $\ell$ -- again using the fact that for the losses in question, ${\cal Q}_{f-f^*}(X_i,Y_i) \gtrsim \ell^{\prime \prime}(Z_i)(f-f^*)^2(X_i)$.

\begin{Theorem} \label{thm:independent-quadratic}
There exist absolute constants $c_1,c_2,c_3$ and $c_4$ for which the following holds. Let $F$ and $W$ be as above. With probability at least $1-2\exp(-c_1\eps^2N)$, for every $f \in F$ that satisfies $\|f-f^*\|_{L_2} \geq 2r_Q$ one has
$$
P_N {\cal Q}_{f-f^*} \geq c_2 \eps \kappa_0^2 \rho(t_1,t_2) \|f-f^*\|_{L_2}^2.
$$
where
$$
t_1=\kappa_1 \|W\|_{L_2} \ \ {\rm and} \ \ t_2= c_3\eps^{-1/2}\|W\|_{L_2} + c_4(\kappa_0+\eps^{-1/2})\|f-f^*\|_{L_2}.
$$
\end{Theorem}
The proof of Theorem \ref{thm:independent-quadratic} is based on several observations leading to accurate information on the `location' of the midpoints $Z_i$ in the lower bound on ${\cal Q}_{f-f^*}(X_i,Y_i)$. For every $(X,Y)$, the corresponding mid-point belongs to interval whose end-points are $(f-f^*)(X)-W$ and $-W$. If $I_f$ is the set of coordinates on which $|(f-f^*)(X_i)|$ is of the order of $\|f-f^*\|_{L_2}$, and since $X$ and $W$ are independent and $W$ is symmetric, then on roughly half of these coordinates the signs of $(f-f^*)(X_i)$ coincide with the signs of $-W_i$. Thus,
$$
|Z_i| \in [|W_i|,|W_i|+|(f-f^*)(X_i)|].
$$
Moreover, by excluding a further, sufficiently small proportion of the coordinates in $I_f$ it follows that $|W_i| \sim \|W\|_{L_2}$, as long as $W$ is not highly concentrated around zero -- which is the reason for \eqref{eq:small-ball-noise}.

The difficulty is in making this argument uniform, in the sense that it should hold for every $f \in F$, rather than for a specific choice of $f$. The first step towards a uniform result is the following lemma.

\begin{Lemma} \label{lemma:signs-simple}
Let $1 \leq k \leq m/40$ and set ${\cal S} \subset \{-1,0,1\}^m$ of cardinality at most $\exp(k)$. For every $s=(s(i))_{i=1}^m \in {\cal S}$ put $I_s=\{i : s(i) \not = 0\}$ and assume that $|I_s| \geq 40 k$. If $(\eps_i)_{i=1}^m$ are independent, symmetric $\{-1,1\}$-valued random variables then with probability at least $1-2\exp(-k)$,
$$
\inf_{s \in {\cal S}} |\{ i \in I_s : {\rm sgn}(s(i))=\eps_i\}| \geq k/3.
$$
\end{Lemma}

\proof For every fixed $s \in {\cal S}$, the event $|\{i: (s(i))_{i \in I_s}=\eps_i\}| \geq \ell$ has the same distribution as $|\{i \in I_s : \eps_i =1\}| \geq \ell$. If $(\delta_i)_{i \in I_s}$ are selectors of mean $1/2$, then
\begin{equation*}
Pr\left(|\{i: (s(i))_{i \in I_s}=\eps_i\}| \geq \ell\right)=Pr \left( \sum_{i=1}^{|I_s|} \delta_i \geq \ell \right)=(*).
\end{equation*}
Applying Bernstein's inequality for $\ell=|I_s|/3$,
\begin{equation*}
(*) \geq 1-Pr\left( \left|\frac{1}{|I_s|}\sum_{i=1}^{|I_s|} \delta_i -\frac{1}{2}\right| \geq \frac{1}{6}\right) \geq  1-2\exp(-|I_s|/20).
\end{equation*}
Therefore, by the union bound,
\begin{align*}
& Pr\left( {\rm for \ every } \ s \in {\cal S}, \ |\{i: (s(i))_{i \in I_s}=\eps_i\}| \geq |I_s|/3 \right) \geq  1-2|{\cal S}| \exp(-|I_s|/20)
\\
\geq & 1-2\exp(-k).
\end{align*}
\endproof

Fix $r$ as in Theorem \ref{thm:small-ball-basic} for the class $H_{f^*}=F-f^*$ and let $\Omega^\prime$ be the event on which its assertion holds.  Using the notation of that theorem, consider $r=2r_Q$ and the set $V_r$. For every $v \in V_r$ and a sample $(X_1,...,X_N) \in \Omega^\prime$, let $I_v = \{ i : \kappa_0 r \leq |v(X_i)| \leq c_1r/\sqrt{\eps} \}$ and set
$$
s_v=\left({\rm sgn}(v(X_i)) \IND_{I_v}(X_i)\right)_{i=1}^N \ \ {\rm and} \ \
{\cal S}=\{s_v : v \in V_r\}.
$$
By Theorem \ref{thm:small-ball-basic}, $Pr(\Omega^\prime) \geq 1-2\exp(-c_2\eps^2 N)$ and on $\Omega^\prime$,
$$
|{\cal S}| \leq \exp(\eps N/1000) \ \ {\rm and} \ \ \min_{v \in V} |I_{v}| \geq \eps N/2.
$$
\begin{Lemma} \label{lemma:mid-points}
Conditioned on $\Omega^\prime$, with probability at least $1-2\exp(-c_0 \eps N)$ with respect to the uniform measure on $\{-1,1\}^N$, the following holds. For every $h \in H_{f^*}$ with $\|h\|_{L_2} \geq r$, there is subset ${\cal I}_h \subset \{1,...,N\}$ of cardinality at least $\eps N/24$, and for every $i \in {\cal I}_h$,
$$
(\kappa_0/2) \|h\|_{L_2} \leq |h(X_i)| \leq c_1 (\kappa_0+1/\sqrt{\eps})\|h\|_{L_2}
 \ \  {\rm and} \ \ {\rm sgn}(h(X_i)) = \eps_i.
$$
\end{Lemma}

\proof Fix $h \in H$ with $\|h\|_{L_2} = r$ and let $\pi h= v \in V_r$ be as in Theorem \ref{thm:small-ball-basic}. Recall that there is a subset $J_h \subset I_v$ consisting of at least $3/4$ of the coordinates of $I_v$, on which
$$
(\kappa_0/2)r \leq |h(X_j)| \leq c_1 (\kappa_0+1/\sqrt{\eps})r \ \ \ {\rm and} \ \ \ {\rm sgn}(h(X_j))={\rm sgn}(v(X_j)).
$$
Applying Lemma \ref{lemma:signs-simple} to the set ${\cal S}=\{s_v : v \in V_r\}$ for $k=\eps N /1000$, and noting that for every $s_v \in {\cal S}$, $|\{i: s_v(i) \not = 0\}| \geq \eps N/2 \geq 40 k$, it follows that with probability at least $1-2\exp(-c_2 \eps N)$ (relative to the uniform measure on $\{-1,1\}^N$), for every $v \in V_r$, $s_v(i)=\eps_i$ on at least $1/3$ of the coordinate of $I_v$.

Since the set $J_h$ contains at least $3/4$ of the coordinates of $I_v$ and $v(X_i)=\eps_i$ on at least a $1/3$ of the coordinates of $I_v$ it follows that on the coordinates that belong to the intersection of these two sets (at least $1/12$ of the coordinates in $I_v$), both conditions hold, as asserted.

Finally, the claim is positive homogeneous and because $H_{f^*}$ is star-shaped around $0$, it holds on the same event when $\|h\|_{L_2} \geq r$.
\endproof

\begin{Corollary} \label{cor:signs-main}
There exist  absolute constants $c_0$ and $c_1$ for which the following holds. Let $F$ and $W$ be as above. With probability at least $1-2\exp(-c_0\eps^2 N)$ with respect to the product measure $(X \otimes W)^N$, for every $f \in F$ with $\|f-f^*\|_{L_2} \geq 2r_Q$ there is a subset ${\cal J}_f \subset \{1,...,N\}$ of cardinality at least $\eps N/100$, and for every $j \in {\cal J}_f$,
\begin{description}
\item{1.} $(\kappa_0/2)\|f-f^*\|_{L_2} \leq |(f-f^*)(X_j)| \leq c_1 (\kappa_0+1/\sqrt{\eps})\|f-f^*\|_{L_2}$,
\item{2.} ${\rm sgn}((f-f^*)(X_i)) = {\rm sgn}(-W)$, and
\item{3.} $\kappa_1 \|W\|_{L_2} \leq |W_j| \leq c_2\|W\|_{L_2}/\sqrt{\eps}$.
\end{description}
\end{Corollary}

\proof Since $W$ is symmetric, it has the same distribution as $\eta |W|$, for a symmetric $\{-1,1\}$-valued random variable $\eta$ that is independent of $|W|$ and of $X$.

If $(W_i)_{i=1}^N=(\eta_i |W_i|)_{i=1}^N$, a direct application of Lemma \ref{lemma:mid-points} shows that with probability at least $1-2\exp(-c_0\eps^2 N)$, if $\|f-f^*\|_{L_2} \geq 2r_Q$, there is a subset ${\cal I}_f \subset \{1,...,N\}$ of cardinality at least $\eps N/24$, and for every $i \in {\cal I}_f$,
$$
(\kappa_0/2)\|f-f^*\|_{L_2} \leq |(f-f^*)(X_j)| \leq  c_1 (\kappa_0+1/\sqrt{\eps})\|f-f^*\|_{L_2}
$$
and
$$
{\rm sgn}((f-f^*)(X_i)) = {\rm sgn}(-\eta_i).
$$
The final component is that for many of the coordinates in ${\cal I}_f$, $|W_i| \sim \|W\|_{L_2}$. Indeed, by excluding the largest and smallest $\eps N/200$ coordinates of $(|W_i|)_{i \in I_f}$, one obtains a subset ${\cal J}_f \subset {\cal I}_f$ of cardinality at least $\eps N/100$, and for every $j \in {\cal J}_f$,
$$
W^*_{N(1-\eps/200)} \leq |W_j| \leq W^*_{\eps N/200},
$$
where $(W_i^*)_{i=1}^N$ is the non-increasing rearrangement of $(|W_i|)_{i=1}^N$.

Observe that by Lemma \ref{lemma-monotone-before-0} applied to $\eps^\prime=\eps/200$, with probability at least $1-2\exp(-c_2 N\eps)$,
$$
W^*_{\eps N/200} \leq c_3\eps^{-1/2}\|W\|_{L_2}.
$$
And, since $Pr(|W| \leq \kappa_1 \|W\|_{L_2}) \leq \eps/1000$, a simple application of a binomial estimate shows that with probability at least $1-2\exp(-c_4 N \eps)$, there are at most $\eps N/200$ $W_i$'s that satisfy $|W_i| < \kappa_1 \|W\|_{L_2}$. Therefore, on that event,
$$
W^*_{(1-\eps/200)N} \geq \kappa_1 \|W\|_{L_2},
$$
completing the proof.
\endproof

\noindent{\bf Proof of Theorem \ref{thm:independent-quadratic}.} Since $\ell$ is convex, ${\cal Q}_{f-f^*}$ is nonnegative. Consider the event from Corollary \ref{cor:signs-main}, and given $f \in F$ for which $\|f-f^*\|_{L_2} \geq 2r_Q$ let ${\cal J}_f \subset \{1,...,N\}$ be the set of coordinates as above. Hence, for every $j \in J_f$,
$$
(\kappa_0/2) \|f-f^*\|_{L_2} \leq |(f-f^*)|(X_j) \leq c_1(\kappa_0+ 1\sqrt{\eps}) \|f-f^*\|_{L_2}.
$$
Moreover, if $j \in J_f$, $-W_j$ and $(f-f^*)(X_j)$ share the same sign, and without loss of generality one may assume that both are positive. Thus, the mid-point $Z_j$ belongs to the interval whose end-points are $t_1=\kappa_1 \|W\|_{L_2}$ and
$t_2=c_1(\kappa_0+ 1\sqrt{\eps})\|f-f^*\|_{L_2}+c_2\|W\|_{L_2}/\sqrt{\eps}$, implying that
\begin{equation*}
P_N {\cal Q}_{f-f^*} \geq \frac{1}{N}\sum_{j \in {\cal J}_f} \ell^{\prime \prime}(Z_i) (f-f^*)^2(X_i)
\geq c_3 \eps \rho(t_1,t_2) \kappa_0^2 \|f-f^*\|_{L_2}^2.
\end{equation*}
\endproof

Next, consider the general noise model, in which $\xi=f^*(X)-Y$ need not be independent of $X$, nor does it necessarily satisfy a small-ball condition.

Observe that the only place in the proof above in which the assumption that $\xi$ and $X$ are independent has been used, was to find a large subset of $\{1,...,N\}$ on which $(f-f^*)(X_i)$ and $\xi_i$ share the same sign. Also, the small-ball assumption on the noise is only used to show that many of the $|\xi_i|$'s are sufficiently large -- of the order of $\|\xi\|_{L_2}$. Both components are not needed if one wishes to show that for a proportional number of coordinates, $|Z_i| \leq c(\kappa_0,\eps)(\|\xi\|_{L_2}+\|f-f^*\|_{L_2})$.

Indeed, it is straightforward to verify that with high probability, if $\|f-f^*\|_{L_2} \geq 2r_Q$, there is a subset of $\{1,...,N\}$ of cardinality at least $\eps N/100$ on which
$$
|Z_i| \lesssim (\kappa_0 + 1/\sqrt{\eps}) \cdot (\|f-f^*\|_{L_2}+\|\xi\|_{L_2}).
$$
Formally one has:
\begin{Theorem} \label{thm:general-loss-general-noise}
There exist absolute constants $c_0,c_1$ and $c_2$ for which the following holds.
Let $F$ be as above, set $Y \in L_2$ and put $\xi=f^*(X)-Y$. Then, with probability at least $1-2\exp(-c_0\eps^2N)$, for every $f \in F$ with $\|f-f^*\|_{L_2} \geq 2r_Q$,
$$
P_N {\cal Q}_{f-f^*} \geq c_1\eps \kappa_0^2 \rho(0,t) \|f-f^*\|_{L_2}^2,
$$
for $t=c_2(\kappa_0+\eps^{-1/2})\cdot(\|f-f^*\|_{L_2}+\|\xi\|_{L_2})$.
\end{Theorem}
\begin{Remark}
The assumption in Theorem \ref{thm:independent-quadratic} that the noise is independent of $X$ allows one to obtain a positive lower bound on $t_1$ -- of the order of the variance $\|W\|_{L_2}$. This is required when the loss function is not strongly convex in a large enough neighbourhood of zero (for example, when $\ell(t)=t^p$ for $p>2$).

When $F$ is bounded in $L_2$, one may use the trivial bound $\|f-f^*\|_{L_2} \leq 2d_F(L_2)$ and replace $t_2$ by $c(\eps,\kappa_0)(\|\xi\|_{L_2} + d_F(L_2))$. This is of little importance when $\ell^{\prime \prime}$ decreases slowly, but it is highly significant when, for example, it has a compact support.

Consider, for example, the Huber loss with parameter $\gamma$. If $\gamma \sim \|\xi\|_{L_2} + d_F(L_2)$ then $\rho(0,t_2)=1$, but as stated, for a smaller value of $\gamma$, $\rho(0,t)=0$ -- leading to a useless estimate on the quadratic component.

It turns out that one may improve Theorem \ref{thm:general-loss-general-noise} dramatically by ruling-out functions in $F$ for which $\|f-f^*\|_{L_2}$ is significantly larger than $\|\xi\|_{L_2}$ as potential empirical minimizers, implying that $t$ can be taken to be $t=c(\kappa_0,\eps)\|\xi\|_{L_2}$. We will present this preliminary exclusion argument in Section \ref{sec:preliminary-exclusion}.
\end{Remark}

\section{Error estimates and oracle inequalities} \label{sec:orcle}
We next turn to the multiplier component of the process, defined by $f \to \frac{1}{N}\sum_{i=1}^N \ell^\prime(\xi_i)(f-f^*)(X_i)$.

\subsection{Multiplier complexity} \label{sec:multi-complexity}
 Let us define a complexity term that may be used to control the multiplier process, and which is similar to the one used in \cite{Shahar-COLT}.

\begin{Definition} \label{def:complexity-linear}
Given a loss function $\ell$, let $\phi^\ell_N(r)$ be the random function
$$
\phi_N^\ell(r)=\frac{1}{\sqrt{N}} \sup_{\{f \in F: \|f-f^*\|_{L_2} \leq r\}} \left|\sum_{i=1}^N \eps_i \ell^\prime(\xi_i)(f-f^*)(X_i)\right|
$$
and set
$$
r_{M}^\prime(\kappa,\delta)=\inf\left\{ r > 0 : Pr \left(\phi_N^\ell(r) \leq  r^2 \kappa \sqrt{N} \right) \geq 1-\delta \right\}.
$$
Recall that $H_{f^*}=F-f^*$, put
$$
r_0(\kappa) = \inf \left\{r :  \sup_{h \in H_{f^*} \cap rD} \|\ell^\prime(\xi)h(X)\|_{L_2} \leq \sqrt{N} \kappa r^2/4 \right\}
$$
and let
$$
r_M(\kappa,\delta)=r_M^\prime(\kappa,\delta) + r_0(\kappa).
$$
\end{Definition}

The function $\phi_N^\ell(r)$ and the definition of $r_M$ arise naturally in a symmetrization argument, that
\begin{align} \label{eq:symm-in-multiplier}
& Pr\left(\sup_{h \in H_{f^*} \cap r D} \left|\frac{1}{N}\sum_{i=1}^N \ell^\prime(\xi_i)h(X_i) - \E \ell^\prime(\xi)h(X)\right| > x \right)
\nonumber
\\
\leq & 2 Pr \left(\sup_{h \in H_{f^*} \cap r D} \left|\frac{1}{N}\sum_{i=1}^N \eps_i \ell^\prime(\xi_i)h(X_i)\right|> \frac{x}{4}\right),
\end{align}
provided that $x \geq 4N^{-1/2}\sup_{h \in H_{f^*} \cap rD} \|\ell^\prime(\xi)h(X)\|_{L_2}$.
\begin{Lemma} \label{lemma:multi-and-zero-level-sym}
If $F$ is a convex class of functions and $r =2 r_M(\kappa/4,\delta/2)$, then with probability at least $1-\delta$, for every $f \in F$ satisfying $\|f-f^*\|_{L_2} \geq r$, one has
$$
\left|\frac{1}{N}\sum_{i=1}^N \ell^\prime(\xi_i)(f-f^*)(X_i) - \E \ell^\prime(\xi)(f-f^*)(X) \right| \leq \kappa \max\{\|f-f^*\|_{L_2}^2,r^2\}.
$$
\end{Lemma}
\proof
Since $F$ is convex, $H_{f^*}=F-f^*$ is star-shaped around $0$. Therefore, as $r \geq r_0$, it follows that $4N^{-1/2}\sup_{h \in H_{f^*} \cap rD} \|\ell^\prime(\xi)h(X)\|_{L_2} \leq \kappa r^2$, and by \eqref{eq:symm-in-multiplier} for $x=\kappa r^2$,
\begin{equation*}
Pr\left(\sup_{h \in H_{f^*} \cap r D} \left|\frac{1}{N}\sum_{i=1}^N \ell^\prime(\xi_i)h(X_i) - \E \ell^\prime(\xi)h(X)\right| > \kappa r^2 \right)
\leq 2 Pr \left( \phi_N^\ell> \frac{ \kappa r^2}{4}\right) \leq \delta,
\end{equation*}
because $r>r_M^\prime(\kappa/4,\delta/2)$.

Using, once again, that $H_{f^*}$ is star-shaped around $0$, if
$\|f-f^*\|_{L_2} \geq r$ then $r(f-f^*)/\|f-f^*\|_{L_2} \in H_{f^*} \cap r S(L_2)$. Thus,
$$
\left|\frac{1}{N}\sum_{i=1}^N \ell^\prime(\xi_i)(f-f^*)(X_i) - \E \ell^\prime(\xi)(f-f^*)(X) \right| \leq \kappa \|f-f^*\|_{L_2}^2.
$$
\endproof

The function $\phi_N^\ell(r)$ is a natural complexity parameter -- it is the `weighted width' of the coordinate projection of $H_{f^*} \cap rD$ in a direction selected from the combinatorial cube -- and scaled according to the `noise multipliers' $(\ell^{\prime}(\xi_i))_{i=1}^N=(f^*(X_i)-Y_i)_{i=1}^N$.

The more standard counterparts of $\phi_N^\ell(r)$, appearing in Theorem \ref{thm:BBM} and in similar results of that flavour, is the random variable
$$
\sup_{f \in F \cap rD_{f^*}} \left|\frac{1}{N}\sum_{i=1}^N \eps_i (f-f^*)(X_i)\right|,
$$
 its conditional expectation -- the so-called {\it Rademacher average}
$$
\E_\eps \left(\sup_{f \in F \cap rD_{f^*}} \left|\frac{1}{N}\sum_{i=1}^N \eps_i (f-f^*)(X_i)\right| \ \big{|} (X_i)_{i=1}^N\right),
$$
and its expectation with respect to both $(\eps_i)_{i=1}^N \otimes (X_i)_{i=1}^N$. Those represent the width or average width relative to a generic noise model, given by $(\eps_1,...,\eps_N)$ for the coordinate projection
$$
P_\sigma (H_{f^*} \cap r D) = \left\{\left((f-f^*)(X_i)\right)_{i=1}^N  : f \in F, \ \|f-f^*\|_{L_2} \leq r \right\}
$$
endowed by the sample $X_1,...,X_N$.

Observe that $\phi_N^\ell$ is inherently superior to the generic complexity term, not only because it does take into account the nature of the noise, something that the generic noise term disregards completely, but also because the multipliers $\ell^\prime(\xi)$ can be trivially removed in the bounded case by applying a contraction argument, thus reverting to the same complexity parameter used in Theorem \ref{thm:BBM}.

\begin{Remark}
It is straightforward to verify that when $F$ consists of heavy-tailed random variables or if $Y$ is a heavy-tailed random variable then the random sets
$$
\left\{ \left(\ell^\prime(\xi_i) (f-f^*)(X_i) \right)_{i=1}^N \ : \ f \in F, \ \|f-f^*\|_{L_2} \leq r \right\}
$$
are weakly bounded, with the unfortunate byproduct that $\phi^\ell_N$ may exhibit rather poor concentration around its conditional mean or true mean. This is why $r_M$ is defined using $\phi_N^\ell$ rather than its expectations, and unlike the bounded case or the subgaussian one, there might be a substantial gap between the fixed point defined using $\phi_N^\ell$ and the one defined using its mean.
\end{Remark}

Combining the bounds on the quadratic and multiplier terms with Theorem \ref{thm:estimation-intro} and Theorem \ref{thm:prediction-intro}, one has the following:
\begin{Theorem} \label{thm:main-combniation-trivial}
For every $\kappa_0$ and $0<\eps<1$ there exist constants $c_0$, $c_1$, $c_2$ and $c_3$ that depend only on $\kappa_0$ and $\eps$, and an absolute constant $c_4$ for which the following holds.
Let $F$ be a convex class of functions and assume that $F-F$ satisfies the small-ball condition with constants $\kappa_0$ and $\eps$. Set $t_1=0$, $t_2=c_0(\kappa_0,\eps)(\|\xi\|_{L_2}+d_F(L_2))$, $\zeta_1=c_1(\kappa_0,\eps)$ and $\zeta_2=c_2(\kappa_0,\eps)$. Put $\theta=c_3(\kappa_0,\eps) \rho(t_1,t_2)$. Then,

\begin{description}
\item{$\bullet$} With probability at least $1-\delta-2\exp(-c_4N\eps^2)$,
$$
\|\hat{f}-f^*\|_{L_2} \leq 2\max\{r_Q(\zeta_1,\zeta_2),r_M(\theta/16,\delta/2)\}.
$$
\item{$\bullet$}
If $\ell$ satisfies Assumption \ref{ass:quadratic-for-oracle} with a constant $\beta$ then with the same probability estimate,
$$
\E {\cal L}_{\hat{f}} \leq 2(\theta+\beta)\max\{r_Q(\zeta_1,\zeta_2),r_M(\theta/16,\delta/2)\}.
$$
\end{description}
\end{Theorem}

\proof By Theorem \ref{thm:general-loss-general-noise}, there is an absolute constant $c_0$ and an event of probability at least $1-2\exp(-c_0\eps^2N)$, on which, if $\|f-f^*\|_{L_2} \geq 2r_Q$ then
$$
P_N {\cal Q}_{f-f^*} \geq \theta \|f-f^*\|_{L_2}^2.
$$
And, by Lemma \ref{lemma:multi-and-zero-level-sym}, on an event of probability at least $1-\delta$, if $\|f-f^*\|_{L_2} \geq 2r_M(\theta/16,\delta/2)$, then
$$
|P_N {\cal M}_{f-f^*} | \leq (\theta/4) \|f-f^*\|_{L_2}^2.
$$
Using the notation of Theorem \ref{thm:estimation-intro} and of Theorem \ref{thm:prediction-intro}, the first event is ${\cal B}$ and the second in ${\cal A}$, and the claim follow.
\endproof

Theorem \ref{thm:main-combniation-trivial} is close to the estimates one would like to establish, with one significant step still missing: $t_2$ is not of the order of $\|\xi\|_{L_2}$ but can be much larger. This is of little significance in the strongly convex case, though for a more general loss it requires an additional argument, which is presented in the next section.

\subsection{Proofs of the main results} \label{sec:preliminary-exclusion}
Let us begin by showing that one may improve the choice of $t_2 =c(\kappa_0,\eps) (\|\xi\|_{L_2}+d_F(L_2))$ to the potentially much better $2c(\kappa_0,\eps)\|\xi\|_{L_2}$. To that end, we will show that with high probability, the empirical minimizer does not belong to the set
$$
\left\{f \in F : \|f-f^*\|_{L_2} \geq \max\left\{\|\xi\|_{L_2},2r_Q\right\}\right\}.
$$
Therefore, the study of ERM may be reduced to the set $F \cap \max\{\|\xi\|_{L_2},2r_Q\} D_{f^*}$, and in which case, Theorem \ref{thm:main-combniation-trivial} may be used directly, as the diameter of the class in question is $\sim \max\{\|\xi\|_{L_2},r_Q\}$.

Recall that
\begin{equation} \label{eq:quad-rep}
{\cal Q}_{f-f^*}= \int_\xi^{\xi+(f-f^*)(X)} (\ell^\prime(w)-\ell^\prime(\xi))dw.
\end{equation}
Using Theorem \ref{thm:general-loss-general-noise}, there are absolute constants $c_0$, $c_1$ and $c_2$ for which, with probability at least $1-2\exp(-c_0\eps^2N)$, if $\|f-f^*\|_{L_2} \geq 2r_Q$, then
\begin{equation} \label{eq:pre-exec-1}
P_N {\cal Q}_{f-f^*} \geq c_1\eps \kappa_0^2 \rho(0,t) \|f-f^*\|_{L_2}^2,
\end{equation}
where $t=c_2(\kappa_0+\eps^{-1/2})\cdot(\|f-f^*\|_{L_2}+\|\xi\|_{L_2})$.
\vskip0.5cm
Let $\theta=c_1\eps \kappa_0^2 \rho(0,t)$ for $t=2c_2(\kappa_0+\eps^{-1/2})\max\{\|\xi\|_{L_2},r_Q\}$ and assume further that
$$
r_M(\theta/16,\delta/2) \leq \max\{\|\xi\|_{L_2},2r_Q\}.
$$

\begin{Theorem} \label{thm:preliminary-exclusion}
On an event of probability at least $1-\delta-2\exp(-c_0N\eps^2)$,
$$
\|\hat{f}-f^*\|_{L_2} \leq \max\{\|\xi\|_{L_2},2r_Q, 2r_M(\theta/16,\delta/2)\}.
$$
\end{Theorem}

The proof of Theorem \ref{thm:preliminary-exclusion} is based on several observations.
\vskip0.5cm
Note that if $h \in F$ and $\|h-f^*\|_{L_2} > R$, there is some $\lambda > 1$ and $f \in F$ for which $\|f-f^*\|_{L_2} = R$ and $\lambda (f-f^*) = (h-f^*)$. Indeed, set $\lambda = \|h-f^*\|_{L_2}/R>1$ and put $f=h/\lambda + (1-1/\lambda)f^*$; by convexity, $f \in F$. Hence, for every $R>0$,
\begin{align} \label{eq:incluion-pre-exclusion}
& \{h-f^* : h \in F \ \ \|h-f^*\|_{L_2} \geq R \} \nonumber
\\
\subset & \{\lambda (f-f^*) : \lambda \geq 1, \ f \in F, \ \|f-f^*\|_{L_2} = R \}.
\end{align}

\begin{Lemma} \label{lemma:quad-pre-exclusion}
On the event on which \eqref{eq:pre-exec-1} holds, if $\|f-f^*\|_{L_2} = \max\{\|\xi\|_{L_2},2r_Q\}$ and $\lambda \geq 1$ then
$$
P_N {\cal Q}_{\lambda (f-f^*)} \geq \lfloor \lambda \rfloor \theta \max\{\|\xi\|_{L_2}^2,4r_Q^2\}.
$$
\end{Lemma}

\proof
Fix $a,x \in \R$ and observe that for every $\lambda \geq 1$,
\begin{equation} \label{eq:integral-observation}
\int_a^{a+\lambda x} (\ell^\prime(w)-\ell^\prime(a) )dw \geq \lfloor \lambda \rfloor \int_a^{a+x} (\ell^\prime(w)-\ell^\prime(a) )dw.
\end{equation}
To see this, let $x>0$ and write
$$
\int_a^{a+\lambda x} (\ell^\prime(w)-\ell^\prime(a) )dw = \sum_{j=0}^{\lfloor \lambda \rfloor-1} \int_{a + jx}^{a+(j+1)x} (\ell^\prime(w)-\ell^\prime(a) )dw + \int_{a + \lfloor \lambda \rfloor}^{a+\lambda x} (\ell^\prime(w)-\ell^\prime(a) )dw.
$$
Since $\ell^\prime(w)-\ell^\prime(a)$ is an increasing function in $w$, the first term in the sum is the smallest and
$$
\int_a^{a+\lambda x} (\ell^\prime(w)-\ell^\prime(a) )dw \geq \lfloor \lambda \rfloor \int_a^{a+x} (\ell^\prime(w)-\ell^\prime(a) )dw.
$$
The case when $x<0$ is equally simple.

When \eqref{eq:integral-observation} is applied to \eqref{eq:quad-rep}, it follows that pointwise,
$$
\int_\xi^{\xi+\lambda(f-f^*)(X)}
\left(\ell^\prime(w)-\ell^\prime(\xi)\right)dw={\cal Q}_{\lambda (f-f^*)} \geq \lfloor \lambda \rfloor {\cal Q}_{f-f^*},
$$
and by the lower bound on $P_N{\cal Q}_{f-f^*}$ the claim follows.
\endproof

\noindent {\bf Proof of Theorem \ref{thm:preliminary-exclusion}.}
Recall that $r_M(\theta/16,\delta/2) \leq \max\{\|\xi\|_{L_2},2r_Q\}$; hence, with probability at least $1-\delta$, if $\|f-f^*\|_{L_2} \leq \max\{\|\xi\|_{L_2},2r_Q\}$ then
$$
| P_N {\cal M}_{f-f^*} | = \left|\frac{1}{N}\sum_{i=1}^N \ell^\prime(\xi_i)(f-f^*)(X_i)\right| \leq (\theta/4) \max\{\|\xi\|_{L_2}^2,4r_Q^2\}.
$$
Since $P_N {\cal M}_{f-f^*}$ is linear in $f-f^*$, it follows that for every $\lambda \geq 1$,
\begin{align*}
|P_N {\cal M}_{\lambda (f-f^*)}|= & \left|\frac{1}{N}\sum_{i=1}^N \ell^\prime(\xi_i)\lambda(f-f^*)(X_i)\right|  = \lambda | P_N {\cal M}_{f-f^*} |
\\
\leq & \lambda (\theta/4) \max\{\|\xi\|_{L_2}^2,4r_Q^2\}.
\end{align*}
Combining this with the lower bound on $P_N {\cal Q}_{f-f^*}$ shows that with probability at least $1-\delta-2\exp(-c_0\eps^2N)$, if $\|f-f^*\|_{L_2} = \max\{\|\xi\|_{L_2},2r_Q\}$ and $\lambda \geq 1$ then
$$
P_N {\cal Q}_{\lambda (f-f^*)} - |P_N {\cal M}_{\lambda (f-f^*)}| \geq \lambda(\theta  /2)\max\{\|\xi\|_{L_2}^2,4r_Q^2\} >0.
$$
Thus, by \eqref{eq:incluion-pre-exclusion}, on that event the empirical minimizer belongs to the set
$$
F \cap \max\{\|\xi\|_{L_2},2r_Q\} D_{f^*}.
$$
\endproof

Now we are finally ready to formulate and prove the main results of the article.
\begin{Theorem} \label{thm:main-combniation}
For every $\kappa_0$ and $0<\eps<1$ there exist constants $c_0$, $c_1$, $c_2$ and $c_3$ that depend only on $\kappa_0$ and $\eps$, and an absolute constant $c_4$ for which the following holds.

Let $F$ be a convex class of functions and assume that $F-F$ satisfies the small-ball condition with constants $\kappa_0$ and $\eps$. Set $t_1=0$ and $t_2=c_0(\eps,\kappa_0)\|\xi\|_{L_2}$, $\zeta_1=c_1(\eps,\kappa_0)$ and $\zeta_2=c_2(\eps,\kappa_0)$. Put $\theta=c_3(\eps,\kappa_0) \rho(t_1,t_2)$.

If $r_M(\theta/16,\delta/2) \leq \max\{\|\xi\|_{L_2},2r_Q(\zeta_1,\zeta_2)\}$, then with probability at least $1-\delta-2\exp(-c_4N\eps^2)$,
\begin{description}
\item{$\bullet$} $\|\hat{f}-f^*\|_{L_2} \leq 2\max\{r_Q(\zeta_1,\zeta_2),r_M(\theta/16,\delta/2)\}$.
\item{$\bullet$}
If $\ell$ satisfies Assumption \ref{ass:quadratic-for-oracle} with a constant $\beta$ then with the same probability estimate,
$$
\E {\cal L}_{\hat{f}} \leq 2(\theta+\beta)\max\{r_Q(\zeta_1,\zeta_2),r_M(\theta/16,\delta/2)\}.
$$
\item{$\bullet$} If $\xi$ is independent of $X$ and satisfies a small-ball condition with constants $\kappa_1$ and $\eps$, one may take $t_1=c_5 \kappa_1 \|\xi\|_{L_2}$ for a constant $c_5=c_5(\eps)$, and the two assertions described above hold as well.
\end{description}
\end{Theorem}
\proof
By the preliminary exclusion argument of Theorem \ref{thm:preliminary-exclusion}, with probability at least $1-\delta-2\exp(-c_0N\eps^2)$, $\|\hat{f}-f^*\|_{L_2} \leq \max\{\|\xi\|_{L_2},2r_Q\}$. If $\|\xi\|_{L_2} \leq 2r_Q$ then Theorem \ref{thm:preliminary-exclusion} suffices to prove the claim. Otherwise, the claim follows by Theorem \ref{thm:main-combniation-trivial}, applied to the class $F \cap \|\xi\|_{L_2}D_{f^*}$.
\endproof
The second main result deals with the case in which $\ell$ is strongly convex in a neighbourhood of zero.
\begin{Theorem} \label{thm:main-combniation-strictly-convex}
For every $\kappa_0$ and $0<\eps<1$ there exist constants $c_0$, $c_1$, $c_2$ and $c_3$ that depend only on $\kappa_0$ and $\eps$, and an absolute constant $c_4$ for which the following holds.

Assume that $\ell$ is strongly convex in the interval $[-\gamma,\gamma]$ with a constant $\kappa_2$ and that $\|\xi\|_{L_2} \leq c_0\gamma$.

Assume further that $F$ is a convex class of functions and that $F-F$ satisfies the small-ball condition with constants $\kappa_0$ and $\eps$. Set $\zeta_1=c_1(\kappa_0,\eps)$, $\zeta_2=c_2(\kappa_0,\eps)$ and $\theta=c_3(\kappa_0,\eps) \kappa_2$.

If $r_M(\theta/16,\delta/2) \leq \gamma$, then with probability at least $1-\delta-2\exp(-c_4N\eps^2)$,
\begin{description}
\item{$\bullet$} $\|\hat{f}-f^*\|_{L_2} \leq 2\max\{r_Q(\zeta_1,\zeta_2),r_M(\theta/16,\delta/2)\}$.
\item{$\bullet$}
If $\ell$ satisfies Assumption \ref{ass:quadratic-for-oracle} with a constant $\beta$ then with the same probability estimate,
$$
\E {\cal L}_{\hat{f}} \leq 2(\theta+\beta)\max\{r_Q(\zeta_1,\zeta_2),r_M(\theta/16,\delta/2)\}.
$$
\end{description}
\end{Theorem}

The proof of Theorem \ref{thm:main-combniation-strictly-convex} is almost identical to that of Theorem \ref{thm:main-combniation}, with one difference: instead of considering the preliminary exclusion argument of Theorem \ref{thm:preliminary-exclusion} at the level $\sim \max\{\|\xi\|_{L_2},2r_Q\}$, one performs preliminary exclusion at the level $\gamma$, and with an identical proof. The rest of the argument remains unchanged and we shall omit its details.

\vskip0.5cm

At this point, let us return to the rather detailed `wish list' that has been outlined in the introduction regarding the parameters governing prediction and estimation problems and see where we stand.

Theorem \ref{thm:main-combniation} and Theorem \ref{thm:main-combniation-strictly-convex} lead to bounds on ${\cal E}_p$ and ${\cal E}_e$ without assuming that the class consists of uniformly bounded or subgaussian functions, nor that the target is even in $L_p$ for some $p>2$. And, under minor smoothness assumption, $\ell$ need not be a Lipschitz function. Thus, all the restrictions of Theorem \ref{thm:BBM} and Theorem \ref{thm:LM-ERM} have been successfully bypassed.

As for the complexity parameters involved, $r_Q$ is indeed an intrinsic parameter of the class $F$ and has nothing to do with the choice of the loss or with the target. It does measure (with the very high probability of $1-2\exp(-c\eps^2N)$), the $L_2$ diameter of the version space of $F$ associated with $f^*$, and thus corresponds to the solution of the noise-free problem.

The noise and loss influence the problem in two places. In the quadratic component, the loss is calibrated to fit the noise level if it is strongly convex in the interval $[0,c_1(\kappa_0,\eps)\|\xi\|_{L_2}]$, or, when the noise is independent, it suffices that the loss is strongly convex in the smaller interval $[c_2(\kappa_1,\eps)\|\xi\|_{L_2},c_1(\kappa_0,\eps)\|\xi\|_{L_2}]$. The strict convexity constant in the interval also fixes the level $\theta$ appearing in the multiplier component.

The main impact of the loss and the noise is seen in the multiplier component, and thus in the external complexity parameter $r_M$.

Indeed, while the quadratic component and the level $\sim \theta$ of a given class will be exactly the same for any loss that has the same strict convexity constant in the interval $[0,c_1(\kappa_0,\eps)\|\xi\|_{L_2}]$, the difference between losses is coded in $r_M$. And, as expected, the interaction between the class, the noise and the loss is captured by a single parameter: the correlation (or width) of a random projection of the localized class with the random vector $(\ell^\prime(\xi_i))_{i=1}^N$.

Therefore, the `wish list' is satisfied in full: without any boundedness assumptions and for a rather general loss function, prediction and estimation problems exhibit the expected two-regime behaviour: a `low-noise' regime captured by an intrinsic parameter and a `high-noise' regime by an external one. The exact nature of the loss and noise determines the external parameter {\it only} through the multiplier $(\ell^\prime(\xi_i))$, and this random vector also determines where the phase transition between the high-noise regime, in which the external parameter $r_M$ is dominant, and the low-noise one, in which the intrinsic parameter $r_Q$ is dominant, takes place.

\vskip0.5cm

The one remaining issue still left open is that a wise choice of the loss may be used to negate the effects of outliers. This will be explored in the next section.

\section{Loss functions and the removal of outliers} \label{sec:pre-examples}
Having filled the list of properties one would like to see in a general prediction/estimation theory, it is interesting to note that as a byproduct, one is given a way of addressing the problem of outliers through the choice of the loss.

Damaging outliers appear when sample points are far from where one would like them to be, {\it and} the loss assigns a large value to those points. This combination means that outliers actually have a true impact on the empirical mean $P_N {\cal L}_f$ and therefore on the identity of the empirical minimizer.

The reason why outliers are of little concern in problems that feature a strong concentration phenomenon is obvious: no matter what the loss is (as long as it does not grow incredibly quickly) only an insignificant fraction of the sample points fall outside the `right area', and thus their effect is negligible.

The situation is different when either the class consists of heavy-tailed functions or when the noise is heavy tailed. In such cases, a more significant fraction of a typical sample falls in a potentially misleading location. If the effect is amplified by a fast-growing loss, outliers become a problem that has to be contended with. This problem may be resolved only by ensuring that the impact of the loss is not overwhelming outside the `expected area' of $[-c\|\xi\|_{L_2},c\|\xi\|_{L_2}]$, which already hints towards the `right choice' of a loss.

As mentioned above, as long as $\ell$ is strongly convex in $[0,c_1(\kappa_0,\eps)\|\xi\|_{L_2}]$ (or in the smaller interval when the noise is independent) the effect of the loss and of the noise is coded in the vector $(\ell^\prime(\xi_i))_{i=1}^N$. While for the squared loss this vector is likely to be large if $\xi$ is heavy-tailed, losses that grow almost linearly in $[a,\infty)$ lead to bounded multipliers, seeing that $|\ell^\prime(\xi)| \lesssim |\ell^\prime(a)|$, and therefore to a smaller multiplier component.

To better explain this observation, we will focus on the three losses mentioned earlier: the squared loss, the logistic loss and the Huber loss.

\begin{description}
\item{$\bullet$} The squared loss is the canonical example of a strongly convex loss with a bounded second derivative; thus it fits both the estimation and the prediction schemes. However, it is susceptible to the problem of outliers because it continues to grow rather rapidly.

\item{$\bullet$} The logistic loss exhibits a strongly convex behaviour in any bounded interval, but with a constant that decreases exponentially quickly to zero with the length of the interval, because its growth becomes close to linear for large values.

\item{$\bullet$}   The Huber loss with parameter $\gamma$ is strongly convex in $(-\gamma,\gamma)$ and grows linearly outside that interval.
\end{description}
We will show that all three losses exhibit the two regimes, but are affected in a different way by outliers.

We will first present estimates using the parameters $r_Q$ and $r_M$ and then bound them for an arbitrary convex, $L$-subgaussian class and a heavy-tailed target\footnote{It should be noted that assuming that $F$ is an $L$-subgaussian class is far from the only situation in which $r_M$ and $r_Q$ may be controlled. However, obtaining the necessary bounds on empirical and multiplier processes  using the `global' structure of the indexing class is a nontrivial problem. To keep the length of this article within reason, results in that direction will be deferred to future work.}.

Finally, we will present two examples of classes of linear functionals on $\R^n$ for each loss: when $F$ is $\R^n$ and when $F_{r,n}$ is the hierarchy generated by $rB_1^n$.

\vskip0.5cm

Let $F \subset L_2$ be a closed, convex class of functions and assume that $F-F$ satisfies a small-ball condition with constants $\kappa_0$ and $\eps$. And, as always, the target one wishes to estimate is $Y \in L_q$ for some $q \geq 2$. For the sake of simplicity, we will assume at times that $q=4$, though this is not really needed in all the examples presented below.

\subsection{Some facts on multiplier processes} \label{sec:pre-facts-in-examples}
The following is an upper estimate on multiplier and empirical processes indexed by a class that is $L$-subgaussian -- which is essentially sharp. It improves a similar result from \cite{shahar-psi2} and its proof may be found in \cite{shahar-almo-iso-small-ball}.

\begin{Theorem} \label{thm:multi-subgaussian-examples}
There exists an absolute constant $c_0$ and for every $L>1$ there are constants $c_1$ and $c_2$ that depend only on $L$ and for which the following holds.

Assume that $\Lambda \in L_q$ for some $q>2$ and that $F$ is $L$-subgaussian. Set $k_F=(\E\|G\|_F/d_F(L_2))^2$ and let $N \geq k_F$.
\begin{description}
\item{$\bullet$}
If $u>c_0$ then with probability at least $1-2\exp(-c_1u^2 k_F)$,
$$
\sup_{f \in F} \left|\frac{1}{N}\sum_{i=1}^N f(X_i) - \E  f \right| \leq c_2u \frac{\E\|G\|_F}{\sqrt{N}}.
$$
\item{$\bullet$}
If $u,\beta >c_0$ then with probability at least $1-2\beta^{-q}N^{-((q/2)-1)}-2\exp(-c_1u^2 k_F)$,
$$
\sup_{f \in F} \left|\frac{1}{N}\sum_{i=1}^N \Lambda_i f(X_i) - \E \Lambda f \right| \leq c_2\beta u \|\Lambda\|_{L_q} \frac{\E\|G\|_F}{\sqrt{N}},
$$
\end{description}
\end{Theorem}

\subsection{The squared loss} \label{subsec:squared}

If $\ell(t)=t^2$ then for every $t_1,t_2$, $\rho(t_1,t_2) = 2$. Therefore, by Theorem \ref{thm:main-combniation-strictly-convex} for an arbitrarily large $\gamma$, it follows that
with probability at least $1-\delta-2\exp(-c_1 \eps^2 N)$,
$$
\|\hat{f}-f^*\|_{L_2} \leq \max\{r_Q(\zeta_1,\zeta_2),r_M(c_2/4,\delta/2)\},
$$
for constants $\zeta_1$, $\zeta_2$ and $c_2$ that depend only on $\kappa_0$ and $\eps$.

Clearly, $\|\ell^{\prime \prime}\|_{L_\infty} \leq 2$. Therefore, with the same probability estimate,
$$
\E {\cal L}_{\hat{f}} \leq 2(c_2+1) \max\{r_Q^2(\zeta_1,\zeta_2),r_M^2(c_2/4,\delta/2)\}.
$$
\vskip0.3cm
When $F$ is, in addition, an $L$-subgaussian class, one may identify the parameters $r_M$ and $r_Q$. Recall that $\|f\|_{\psi_2} \sim \sup_{p \geq 2} \|f\|_{L_p}/\sqrt{p}$ and as noted earlier, this suffices to ensure that the small-ball condition holds for $F-F$, and $\kappa_0$ and $\eps$ can be taken to be constants that depend only on $L$.

Since
$$
r_{2,Q}(\zeta_2)=\inf\left\{r>0: \E \sup_{f \in F \cap rD_{f^*}} \left|\frac{1}{\sqrt{N}}\sum_{i=1}^N \eps_i (f-f^*)(X_i) \right| \leq \zeta_2 r \sqrt{N} \right\},
$$
and setting $F_r=\{ f-f^* : f \in F \cap rD_{f^*} \}$, it follows from
Theorem \ref{thm:multi-subgaussian-examples} that
$$
\E \sup_{f \in F \cap rD_{f^*}} \left|\frac{1}{\sqrt{N}}\sum_{i=1}^N \eps_i (f-f^*)(X_i) \right| \leq c_3 \E\|G\|_{F_r}.
$$
Therefore,
$$
r_Q(\zeta_1,\zeta_2) =\inf \left\{r>0: \E\|G\|_{F_r} \leq c_4 \min\{\zeta_1,\zeta_2\} r \sqrt{N} \right\}.
$$
Turning to $r_M$, one has to identify
$$
r_0=\inf\left\{r>0: \sup_{f \in F \cap r D_{f^*}} \|\xi (f-f^*)(X)\|_{L_2} \leq \sqrt{N} r^2 (c_2/16)\right\}
$$
and the `lowest' level $r$ for which
$$
Pr\left(\frac{1}{\sqrt{N}} \sup_{f \in F \cap r D_{f^*}} \left|\sum_{i=1}^N \eps_i \xi_i (f-f^*)(X_i) \right| \leq r^2 (c_2/4) \sqrt{N} \right) \geq 1-\delta.
$$
Since $\|f-f^*\|_{L_4} \leq 2L\|f-f^*\|_{L_2}$, one has that
$$
\|\xi (f-f^*)(X)\|_{L_2} \leq 2L\|\xi\|_{L_4} r \lesssim_L \sqrt{N} r^2,
$$
provided that $r \gtrsim \|\xi\|_{L_4}/\sqrt{N}$.

As for the second term, by Theorem \ref{thm:multi-subgaussian-examples} for $q=4$, it follows that with probability at least $1-2/(\beta^{4}N)-2\exp(-c_3(L)u^2 k_{F_r})$,
$$
\sup_{f \in F_r} \left|\frac{1}{\sqrt{N}}\sum_{i=1}^N \eps_i \xi_i (f-f^*)(X_i)\right| \leq c_4(L)\beta u \|\xi\|_{L_q} \E\|G\|_{F_r}.
$$
Fix $0<\delta<1$. If $k_{F_r} \geq \log(2/\delta)$ one may take $u=c_5(L)$ and if the reverse inequality is satisfied, one may set $u \sim_L (k_{F_r}^{-1}\log(2/\delta))^{1/2}$, leading to a probability estimate of $1-\delta$. Therefore, if
$$
u(r,\delta)=c_6(L) \left(1+k_{F_r}^{-1}\log(2/\delta)\right)^{1/2}
$$
and
$$
\beta \sim \max\left\{\frac{1}{(\delta N)^{1/4}},1\right\},
$$
then with probability at least $1-\delta$,
$$
\sup_{f \in F_r} \left|\frac{1}{\sqrt{N}}\sum_{i=1}^N \eps_i \xi_i (f-f^*)(X_i)\right| \leq c_7(L)\beta u(r,\delta) \|\xi\|_{L_4} \E\|G\|_{F_r},
$$
and
\begin{equation} \label{eq:rm-in-example-squared1}
r_M(c_2/4,\delta/2) \leq \frac{\|\xi\|_{L_4}}{\sqrt{N}} + \inf \{r>0: \E\|G\|_{F_r} \leq c_8(L)  \sqrt{N} \|\xi\|_{L_4}^{-1}(\beta u)^{-1}r^2\}.
\end{equation}
Thus, $r_M(c_2/4,\delta/2)$ dominates $r_Q(\zeta_1,\zeta_2)$
as long as $\|\xi\|_{L_4}$ is not very small.

\vskip0.5cm

As a point of reference, consider the case in which the infimum in \eqref{eq:rm-in-example-squared1} is attained for a value $r$ for which $u(r,\delta)=c(L)$. Therefore,
\begin{equation} \label{eq:rm-in-example-squared2}
r_M(c_2/4,\delta/2) \leq \frac{\|\xi\|_{L_4}}{\sqrt{N}} + \inf \{r>0: \E\|G\|_{F_r} \leq c_9(L)  \sqrt{N} \|\xi\|_{L_4}^{-1}\beta^{-1}r^2\}.
\end{equation}
The difference between \eqref{eq:rm-in-example-squared2} and the analogous estimate in the purely subgaussian case (Theorem \ref{thm:LM-ERM}) is the factor $\beta^{-1}$, which causes a slower rate when the desired confidence level is high. Indeed, if $\delta \ll 1/N$, then $\beta \gg 1$ leading to a larger value of $r_M$ than in the subgaussian case.

The different rate is caused by the outliers one encounters -- it is the price for using the squared loss in a heavy-tailed scenario ($\xi \in L_4$ rather than $\xi \in L_{\psi_2}$) leading to a polynomial dependence on $1/\delta$ rather than the logarithmic one exhibited in a purely subgaussian problem.

\vskip0.5cm
\subsection{The logistic loss} \label{subsec:logit-loss}
It is straightforward to verify that for $t \geq 0$, $\ell^\prime(t)=1-2/(\exp(t)+1)$, and thus $\ell^\prime(t) \leq \min\{2t,1\}$.
Also, $\ell^{\prime \prime}(t)= 2\exp(t)/(\exp(t)+1)^2$, which is a decreasing function on $\R_+$ and is upper-bounded by $1$.

Therefore, $\ell$ satisfies Assumption \ref{ass:quadratic-for-oracle}, and one may verify that $\theta \sim \rho(t_1,t_2) \geq \exp(-c_1\|\xi\|_{L_2})$.

The difference between using the logistic loss and the squared loss is seen in $r_M(\theta/16,\delta/2)$. While the multipliers for the squared loss are independent copies of $\ell^\prime(\xi)=\xi$, for the logistic loss, one has $|\ell^\prime(\xi)| \leq \min\{2|\xi|,1\}$. Hence, the multipliers are effectively truncated at $1$. This almost linear growth of the loss outside $[0,1]$ helps one overcome the issue of outliers and leads to an improved estimate -- a logarithmic dependence in $1/\delta$ rather than the polynomial one exhibited by the squared loss. However, the improved error rate does not come for free: it is not satisfied when $\|\xi\|_{L_4}$ is close to zero.

Indeed, the improved estimate is based on a contraction argument. Since $|\ell^\prime(\xi)|\leq 1$, one has
\begin{align*}
& Pr\left( \sup_{f \in F \cap r D_{f^*}} \left|\frac{1}{N}\sum_{i=1}^N \eps_i \ell^\prime(\xi_i) (f-f^*)(X_i)\right| > t \right)
\\
\leq & 2Pr\left( \sup_{f \in F \cap r D_{f^*}} \left|\frac{1}{N}\sum_{i=1}^N \eps_i (f-f^*)(X_i)\right| > t \right),
\end{align*}
removing any dependence on the multipliers. This is a costly step when $\|\xi\|_{L_4}$ is very small, but a necessary one if the aim is to obtain a logarithmic dependence on $1/\delta$.

Therefore, if $u(r,\delta)$ is as defined above,
$$
r_M(\theta/16,\delta/2) \leq \frac{1}{\sqrt{N}} + \inf \{r>0: \E\|G\|_{F_r} \lesssim \theta u(r,\delta)\sqrt{N} r^2\},
$$
and if $r_M \leq \|\xi\|_{L_2}$, then with probability at least $1-\delta-2\exp(-c_0\eps^2 N)$,
$$
\|\hat{f}-f^*\|_{L_2} \lesssim 2\max\{r_Q,r_M\} \ \ {\rm and} \ \ \E {\cal L}_{\hat{f}} \lesssim \max\{r_Q^2,r_M^2\}.
$$
Again, consider that case in which $\inf \{r>0: \E\|G\|_{F_r} \lesssim \theta u(r,\delta)\sqrt{N} r^2\}$ is attained by a value $r$ for which $u(r,\delta)=c(L)$. Then,
$$
r_M(\theta/16,\delta/2) \leq \frac{1}{\sqrt{N}} + \inf \{r>0: \E\|G\|_{F_r} \lesssim \theta \sqrt{N} r^2\},
$$
leading to a far better result than for the squared loss when $\|\xi\|_{L_4} \sim 1$.

The improved rates occur when $\|\xi\|_{L_4} \sim 1$ simply because the logistic loss is calibrated to perform well at that noise level -- but this is no more than a coincidence. The logistic loss is not calibrated to the true noise level of the problem, and indeed the rates deteriorate when $\|\xi\|_{L_4}$ is either very large or very small.
\vskip0.5cm
\subsection{The Huber loss} \label{subsub-Huber-loss}
Let $r_Q$ be as above for suitable constants $\zeta_1$ and $\zeta_2$. For $\zeta_3=\min\{\zeta_1,\zeta_2\}$ one has
$$
r_Q=\inf \{r>0: \E \|G\|_{F_r} \leq c\zeta_3 r \sqrt{N}\}
$$
for a constant $c=c(L)$. Without loss of generality, one may assume that $c \zeta_3$ is smaller than any fixed constant -- which, will be the constant $c_3=c_3(L)$ defined below.

Let $\ell(t)$ be the Huber loss with parameter $\gamma =c_0(L) \max\{\|\xi\|_{L_2},r_Q\}$, for a constant $c_0$ that will be specified later.

Observe that $|\ell^\prime(t)|=\min\{|t/2|,\gamma\}$, that $\ell^\prime$ is a Lipschitz function with constant $1$ and that Assumption \ref{ass:quadratic-for-oracle} is verified. Moreover, this setup falls within the scope of Theorem \ref{thm:main-combniation-strictly-convex}, for $\theta=c_1(L)$.

Regarding the multiplier component, note that if $\|f-f^*\|_{L_2} \leq r$ then
$$
\|\ell^\prime(\xi)(f-f^*)\|_{L_2} \leq \gamma \|f-f^*\|_{L_2} \leq \gamma r \leq (c_1(L)/16) r^2\sqrt{N},
$$
provided that $r \gtrsim_L \gamma/\sqrt{N}$. A contraction argument shows that
\begin{align*}
& Pr\left( \sup_{f \in F \cap r D_{f^*}} \left|\frac{1}{N}\sum_{i=1}^N \eps_i \ell^\prime(\xi_i) (f-f^*)(X_i)\right| > t \right)
\\
\leq & 2Pr\left(  \sup_{f \in F \cap r D_{f^*}} \left|\frac{1}{N}\sum_{i=1}^N \eps_i (f-f^*)(X_i)\right| > \frac{t}{\gamma} \right).
\end{align*}
Therefore, if $u(r,\delta)$ is as defined above, one has
$$
r_M \leq c_2(L) \left( \frac{\max\{\|\xi\|_{L_2},r_Q\}}{\sqrt{N}} + \inf \{r>0: \gamma \E\|G\|_{F_r} \leq c_1(L)  u(r,\delta)\sqrt{N} r^2\} \right).
$$
Again, let us consider the case in which $\inf \{r>0: \gamma \E\|G\|_{F_r} \leq c_2  u(r,\delta)\sqrt{N} r^2\}$ is attained for a value $r$ for which $u(r,\delta)=c^\prime(L)$.
Hence, setting $c_3=c_1(L)u$,
\begin{equation} \label{eq:r-M-in-Huber-example}
r_M \leq c_2\left( c_0 \frac{\max\{\|\xi\|_{L_2},r_Q\}}{\sqrt{N}} + \inf \{r>0: \gamma \E\|G\|_{F_r} \leq c_3  \sqrt{N} r^2\}\right).
\end{equation}
Observe that if $r_Q \leq \|\xi\|_{L_2}$ then $\gamma= c_0\|\xi\|_{L_2}$. Since $c_3 \geq c\zeta_3$,  $r=\|\xi\|_{L_2}$ belongs to the set $\{r>0: c_0\|\xi\|_{L_2} \E\|G\|_{F_r} \leq c_3  \sqrt{N} r^2\}$ in \eqref{eq:r-M-in-Huber-example}. Therefore,
$$
r_M \leq  c_2\left(1+\frac{c_0}{\sqrt{N}}\right)\|\xi\|_{L_2} \leq c_0\|\xi\|_{L_2}=\gamma
$$
if $N \geq c_4(L)$ and for a well chosen $c_0$. Hence, the assumption of Theorem \ref{thm:main-combniation-strictly-convex} is verified.

Otherwise, $\|\xi\|_{L_2} \leq r_Q$, and in which case, $r_Q$ belongs to the set in \eqref{eq:r-M-in-Huber-example} implying that
$$
r_M \leq c_2\left(1+\frac{c_0}{\sqrt{N}}\right)r_Q \leq \gamma.
$$
By Theorem \ref{thm:main-combniation-strictly-convex},
with probability at least $1-\delta-2\exp(-c_0\eps^2N)$,
$$
\|\hat{f}-f^*\|_{L_2} \lesssim \max\{r_M,r_Q\} \ \ {\rm and} \ \ \E{\cal L}_{\hat{f}} \lesssim \max\{r_M^2,r_Q^2\}.
$$
Thanks to the right choice of $\gamma$ in the Huber loss, giving one the optimal interval of strong convexity $[0, c\max\{\|\xi\|_{L_2},r_Q\}]$ relative to the class and the noise, one obtains a far better estimate than for the squared loss. In fact, ${\cal E}_e$ coincides with the purely subgaussian estimate of Theorem \ref{thm:LM-ERM}, with one obvious improvement -- $\|\xi\|_{L_2}$ replaces $\|\xi\|_{\psi_2}$.

\subsection{Examples}
Next, let us present two concrete examples in which the rates can be computed explicitly, and which show how they are affected by the choice of the loss.

Let $T \subset \R^n$, set $F_T=\{\inr{t,\cdot} : t \in T\}$ and assume that $\mu$ is an isotropic, $L$-subgaussian measure on $\R^n$. Therefore, its covariance structure coincides with the standard $\ell_2^n$ distance on $\R^n$ (isotropicity), and  for every $t \in S^{n-1}$ and every $p \geq 2$, $\|\inr{X,t}\|_{L_p} \leq L\sqrt{p} \|\inr{X,t}\|_{L_2}$ ($L$-subgaussian). In particular, $F-F$ satisfies the small-ball condition with constants that depend only on $L$.

\subsubsection{Example I: $\R^n$ as a class of linear functionals} \label{sec:example-unit-ball}

Let $T=\R^n$. Clearly, for every $r>0$ and any possible $f^* \in F$, $F_r=rB_2^n$, implying that $\E\|G\|_{F_r} \sim r\sqrt{n}$. Therefore, $k_{F_r} \sim \sqrt{n}$ and $\E\|G\|_{F_r} \leq \alpha \sqrt{N} r^2$ when $r \geq \alpha^{-1}\sqrt{n/N}$.
\begin{description}
\item{$\bullet$} {\bf The squared loss.}  It is straightforward to verify that if $N \geq c_1(L)n$, then with probability at least $1-2\exp(-c_2(L)N)$, $r_Q=0$. Also,
    $$
    u(r,\delta)=c_3(L) \left(1+\sqrt{\frac{\log(1/\delta)}{N}}\right).
    $$
Using the definition of $r_M$, it is evident that with probability at least $1-\delta-2\exp(-c_4(L)N)$,
\begin{equation} \label{eq:example-squared-linear-R-n}
\|\hat{f}-f^*\|_{L_2} \lesssim_L \max\left\{\frac{1}{(N\delta)^{1/4}},1\right\}  \cdot \left( 1+\sqrt{\frac{\log(1/\delta)}{n}}\right) \cdot \|\xi\|_{L_4} \sqrt{\frac{n}{N}},
\end{equation}
exhibiting once again that the rate has a polynomial dependence in $1/\delta$.

\item{$\bullet$} {\bf The logistic loss.} Let $t_2 \sim_{L} \|\xi\|_{L_2}$ and therefore, $\theta \sim_L \exp(-c_1\|\xi\|_{L_2})$. One has to take $N \geq c_2  n$ to ensure a nontrivial bound on $r_Q$, and in which case, $r_Q=0$.
Therefore, and in a similar way to the squared loss, with probability at least $1-\delta-2\exp(-c_3(L)N)$
$$
\|\hat{f}-f^*\|_{L_2} \lesssim_L \exp(c_1(L)\|\xi\|_{L_2}) \left(1+\sqrt{\frac{\log(1/\delta)}{n}}\right) \cdot \sqrt{\frac{n}{N}},
$$
which is better than \eqref{eq:example-squared-linear-R-n} in terms of the dependence on $\delta$ when $\|\xi\|_{L_2}$ is of the order of a constant and $\delta \ll 1/N$, but does not scale correctly with $\|\xi\|_{L_2}$ when the norm is either very small or very large. This was to be expected from the `calibration' of the logistic loss, which only fits a constant noise level.

\vskip0.3cm
\item{$\bullet$} {\bf The Huber loss.} As noted above, for a nontrivial bound on $r_Q$ one must take $N \geq c_1(L)n$, and in which case, $r_Q=0$. Fix $\gamma =c_2(L)\max\{\|\xi\|_{L_2},r_Q\}=c_2(L) \|\xi\|_{L_2}$ and $\theta=c_2(L)$. Using the definition of $r_M$ (because $|\ell^\prime(\xi)| \leq \gamma$), it is evident that with probability at least $1-\delta-2\exp(-c_3(L)N)$,
$$
\|\hat{f}-f^*\|_{L_2} \leq c_{4}(L) \left(\sqrt{\frac{\log(1/\delta)}{n}}+1\right) \cdot \|\xi\|_{L_2} \sqrt{\frac{n}{N}}.
$$
This is the optimal estimate for any choice of $\|\xi\|_{L_2}$ and coincides with the optimal rate for the squared loss when $\xi$ is gaussian and independent of $X$ (see, e.g. \cite{LM13}).

The optimal rate is obtained by this choice of the Huber loss because it is calibrated to fit the noise level of the problem and the intrinsic complexity of the class.
\end{description}

\subsubsection{Example II:  $\alpha B_1^n$} \label{sec:ball-l-1-n}
Finally, we will sketch, omitting most of the details, the bounds for the squared loss and for the Huber loss in the persistence problem (see Appendix \ref{app:persistence} for some details on the problem). Roughly put, the question is to bound ${\cal E}_p$ and ${\cal E}_e$ for the class of linear functionals indexed by $T_{\alpha,n}=\alpha B_1^n$.

A sharp lower bound on ${\cal E}_p$ and ${\cal E}_e$ relative to the squared loss for these classes (at least for $\alpha=1$ -- though the modifications required for a general $\alpha$ are minimal) and when the noise is a gaussian variable that is independent of $X$, may be found in \cite{LM13}. We will show here that if one uses a well calibrated Huber loss, one may obtain the optimal bounds - as if $\xi$ were gaussian and independent of $X$, even when $\xi$ is actually a heavy-tailed random variable.

\vskip0.5cm

Since $B_1^n \cap r B_2^n$ is equivalent to ${\rm conv}\left(r \bigcup_{|I|=r^2} B_2^I\right)$ -- the convex hull of the union of all Euclidean balls of radius $r$ that are supported on $r^2$ coordinates, it is standard to verify that
\begin{equation*}
\E\sup_{t \in B_1^n \cap r B_2^n} \sum_{i=1}^n g_it_i \sim
\begin{cases}
\sqrt{\log(enr^2)} & \ \ {\rm if} \ \ r \geq 1/\sqrt{n},
\\
\\
r\sqrt{n} & \ \  {\rm if} \ \ r \leq 1/\sqrt{n}.
\end{cases}
\end{equation*}
Since $\mu$ is an isotropic measure, its covariance structure coincides with the standard inner product in $\R^n$. Therefore,
$$
F_{\alpha,r}=F_\alpha \cap r D=\left\{\inr{t,\cdot} : t \in \alpha B_1^n \cap r B_2^n\right\}.
$$
Recall that $k_{F_{\alpha,r}}^{1/2}=\E\|G\|_{F_{\alpha,r}}/r$. Hence,
\begin{equation*}
k_{F_{\alpha,r}}^{1/2} \sim
\begin{cases}
 \frac{\alpha}{r} \log^{1/2}(en(r/\alpha)^2) & \ \ {\rm if} \ \ r\sqrt{n}
 \geq \alpha,
 \\
 \\
 \\
 \sqrt{n} & \ \ {\rm if} \ \ r\sqrt{n} < \alpha.
\end{cases}
\end{equation*}
Also, since
$$
u(r,\delta) \sim_L (1+k_{F_{r,\alpha}}^{-1} \log(2/\delta))^{1/2},
$$
it is evident that $u(r,\delta) \E \|G\|_{F_{r,\alpha}} \sim_L \E\|G\|_{F_{r,\alpha}} + r \log^{1/2}(2/\delta)$.

Recall that $r_Q(\zeta_1,\zeta_2) \leq \inf\{r : k^{1/2}_{F_r} \leq c_0(L) \sqrt{N} \zeta_3\}$ for $\zeta_3=\min\{\zeta_1,\zeta_2\}$, which is a constant that depends only on $L$. Therefore,
\begin{equation*}
r_Q \lesssim_L
\begin{cases}
\frac{\alpha}{\sqrt{N}} \sqrt{\log(en/N)} & \ \ {\rm if} \ \ N \leq c_1(L)n,
\\
\\
\frac{\alpha}{\sqrt{n}} & \ \ {\rm if} \ \ c_1(L)n \leq N \leq c_2(L)n,
\\
\\
0 & \ \ {\rm if} \ \ N \geq c_2(L)n.
\end{cases}
\end{equation*}

As for the multiplier component, a straightforward yet tedious computation shows that for the squared loss $\beta \sim  \max\{1/(\delta N)^{1/4},1\}$, and
\begin{equation*}
r_M^2 \lesssim_L \beta^2\|\xi\|_{L_4}^2 \frac{\log(2/\delta)}{N} +
\begin{cases}
\frac{\alpha \beta \|\xi\|_{L_4}}{\sqrt{N}} \log^{1/2}\left(\frac{en \beta \|\xi\|_{L_4}}{\alpha \sqrt{N}}\right) & \ \ {\rm if} \ \ \alpha \leq \beta \|\xi\|_{L_4} \frac{n}{\sqrt{N}},
\\
\\
\\
\beta^2 \|\xi\|_{L_4}^2 \frac{n}{N} & \ \ {\rm if} \ \ \alpha \geq \beta \|\xi\|_{L_4} \frac{n}{\sqrt{N}},
\end{cases}
\end{equation*}
leading once again to a polynomial dependence on $1/\delta$.

In contrast, a similar estimate for the Huber loss with parameter $\gamma \sim_L \max\{\|\xi\|_{L_2},r_Q\}$, shows that
\begin{equation*}
r_M^2 \lesssim_L \|\xi\|_{L_2}^2 \frac{\log(2/\delta)}{N} +
\begin{cases}
\gamma \frac{\alpha}{\sqrt{N}} \log^{1/2}\left(\frac{e n \gamma}{\alpha \sqrt{N}}\right) & \ \ {\rm if} \ \ \alpha \leq \frac{n \gamma}{\sqrt{N}},
\\
\\
\\
\gamma^2\frac{n}{N} & \ \ {\rm if} \ \ \alpha \geq \frac{n \gamma}{\sqrt{N}}.
\end{cases}
\end{equation*}
Combining the estimates on $r_Q$ and $r_M$, one may show that for the Huber loss, the estimate of $\|\hat{f}-f^*\|_{L_2} \leq 2 \max\{r_M,r_Q\}$ that holds with probability at least $1-\delta-2\exp(-c(L)N)$ is actually the minimax rate for the persistence problem (see, e.g., \cite{LM13}), when $\xi$ is a gaussian variable that is independent of $X$.

\begin{footnotesize}

\bibliographystyle{plain}

\bibliography{biblio}
\end{footnotesize}

\newpage
\appendix
\section{The Classical method} \label{sec:the-classical-method}
Here, we will present a simple proof of Theorem \ref{thm:BBM} that illustrates the main ideas of the classical method.

\begin{Assumption} \label{as:classical}
Assume that
\begin{description}
\item{1.} The loss $\ell$ is a Lipschitz function with a constant $L$ in $[-2b,2b]$.
\item{2.} The class $F$ consists of functions that are bounded by $b$ in $L_\infty$ and so is the target $Y$.
\item{3.} The excess loss ${\cal L}$ satisfies a Bernstein-type condition: there is a constant $B$ such that for every $f \in F$,
    $$
    \|f-f^*\|_{L_2}^2 \leq B \E {\cal L}_f.
    $$
\end{description}
Without loss of generality, we will assume that $b,B \geq 1$.
\end{Assumption}

Out of these three assumptions, it is straightforward to relax (2), by assuming that the class $F$ has a well behaved envelope function $H(x)=\sup_{f \in {F}} |f(x)|$ that belongs to $L_p$ or to $L_{\psi_\alpha}$. Having said that, it should be noted that such an assumption does not really go beyond the bounded case. An envelope condition restricts the `peaky' part of each function to a fixed area (exactly where the envelope is large), and so it may be controlled by studying a single function, rather than a class of functions. Thus, by applying a simple truncation argument, one reverts to the bounded case.

As noted in the introduction, (1) and (2) are restrictive and somewhat unrealistic assumptions.

As for assumption (3), one may show that it holds if ${F} \subset L_2$ is a convex set and the loss $\ell$ is strongly convex. In that case, the constant $B$ depends only on the strict convexity constant of $\ell$. It also holds even when ${F}$ is not convex, for example, when $\ell$ is the squared loss and $Y=f_0(X)+W$, for some $f_0 \in F$ and a mean-zero random variable $W$ that is independent of $X$.

A more difficult observation is that the same is true for any target $Y$ that is `far away' from the set of functions $Y^\prime$ for which $\E\ell(Y^\prime-f)$ has multiple minimizers in ${F}$. In such a case, the constant $B$ depends on the distance between $Y$ and the set of `bad targets' \cite{MR2426759}.

Observe that by combining (1) and (3), it follows that for every $f \in {F}$,
\begin{align} \label{eq:full-Bernstein}
\E {\cal L}_f^2 = & \E \left(\ell(f(X)-Y)-\ell(f^*(X)-Y)\right)^2 \leq L^2 \E|f-f^*|^2
\nonumber
\\
\leq & B L^2 \E {\cal L}_f,
\end{align}
which is the standard Bernstein condition (see, e.g., \cite{MR2166554,MR2240689}).

Recall that $H_{f^*}=F-f^*$ and that $D$ is the $L_2(\mu)$ unit ball. Put
\begin{align} \label{eq:comp-term-bounded}
\phi_N(r)= & \frac{1}{\sqrt{N}} \sup_{\{f \in F : \|f-f^*\|_{L_2} \leq r\}} \left|\sum_{i=1}^N \eps_i (f-f^*)(X_i) \right| \nonumber
\\
= & \frac{1}{\sqrt{N}} \sup_{\{h \in H_{f^*} \cap r D\}} \left|\sum_{i=1}^N \eps_i h(X_i) \right|,
\end{align}
and
\begin{equation} \label{eq:comp-term-r-avg}
\bar{k}_N(\gamma) = \inf \left\{ r>0 : \E \phi_N(r/L) \leq \gamma r^2 \sqrt{N}\right\},
\end{equation}
where the expectation is taken with respect to both $(\eps_i)_{i=1}^N$ and $(X_i)_{i=1}^N$.

The fact that $F$ is convex comes in handy not only for the Bernstein condition, but also to show that $H_{f^*}$ is star-shaped around $0$, which leads to the following:

\begin{Lemma} \label{lemma:star-1}
If $r>\bar{k}_N(\gamma)$ then $\E \phi_N(r) \leq \gamma r^2 \sqrt{N}$, and if $r<\bar{k}_N(\gamma)$, the reverse inequality holds.
\end{Lemma}

\proof
Fix $\rho_1>0$ for which
$$
\E \phi_N(\rho_1)=\E \sup_{h \in H_{f^*} \cap \rho_1 D} \left|\frac{1}{\sqrt{N}}\sum_{i=1}^N \eps_i h(X_i)\right| \leq \gamma \rho_1^2 \sqrt{N},
$$
and note that if $\rho_2>\rho_1$ and $h \in H_{f^*}$ with $\|h\|_{L_2} = \rho_2$ then $(\rho_1/\rho_2)h \in H_{f^*} \cap \rho_1D$. Given $(\eps_i)_{i=1}^N$ and $(X_i)_{i=1}^N$, assume that $\sup_{h \in H_{f^*} \cap \rho_2D} \left|\sum_{i=1}^N \eps_i h(X_i) \right|$ is attained in $h$ and that $\rho_1 \leq \|h\|_{L_2} \leq \rho_2$. Therefore,
\begin{align*}
& \sup_{h \in H_{f^*} \cap \rho_2 D} \left|\sum_{i=1}^N \eps_i h(X_i)\right|=\sup_{h \in H_{f^*} \cap \rho_2 D} \frac{\|h\|_{L_2}}{\rho_1} \left|\sum_{i=1}^N \eps_i \frac{\rho_1}{\|h\|_{L_2}} h(X_i) \right|
\\
\leq & \frac{\rho_2}{\rho_1} \sup_{u \in H_{f^*} \cap \rho_1 D}\left|\sum_{i=1}^N \eps_i u(X_i) \right|.
\end{align*}
Taking expectations on both sides,
\begin{align*}
& \E \sup_{h \in H_{f^*} \cap \rho_2 D} \left|\frac{1}{\sqrt{N}}\sum_{i=1}^N \eps_i h(X_i) \right|
\leq \frac{\rho_2}{\rho_1} \E \sup_{u \in H_{f^*} \cap \rho_1 D}\left|\frac{1}{\sqrt{N}}\sum_{i=1}^N \eps_i u(X_i) \right|
\\
\leq & \gamma \rho_2 \rho_1 \sqrt{N}\leq \gamma \rho_2^2\sqrt{N}.
\end{align*}
The proof of the second part follows an identical path and is omitted.
\endproof

The proof of Theorem \ref{thm:BBM} relies heavily on Talagrand's concentration inequality for bounded empirical processes, a version of which, due to Bousquet \cite{MR1890640} (see also \cite{BouLugMass13}), is formulated below.
\begin{Theorem} \label{thm:talagrand-conc}
There exist an absolute constant $C$ for which the following holds. Let $H$ be a class of functions and set $\sigma_{H} = \sup_{h \in {H}} \|h\|_{L_2}$ and $b=\sup_{h \in {H}} \|h\|_{L_\infty}$. For every $x>0$, with probability at least $1-2\exp(-x)$,
$$
\sup_{h \in H} \left|\frac{1}{N}\sum_{i=1}^N \eps_i h(X_i) \right| \leq C \left(\E\sup_{h \in H} \left|\frac{1}{N}\sum_{i=1}^N \eps_i h(X_i) \right| + \sigma_{H} \sqrt{\frac{x}{N}}+b\frac{x}{N} \right).
$$

\end{Theorem}

The classes we will be interested in are level sets of $F$, scaled according to the excess risk: let $r=4\max\{\bar{k}_N(\gamma),L\sqrt{B}/\sqrt{N}\}$ and put
$$
F_j = \{ f \in F :  2^{j-1} r^2 \leq \E {\cal L}_f < 2^{j}r^2\}, \ \ {\rm and} \ \ F_0=\{f \in F : \E {\cal L}_f \leq r^2\}.
$$
Since $\ell$ is a Lipschitz function with a constant $L$ and the excess loss satisfies a Bernstein condition with a constant $B$, then by the first and third parts of Assumption \ref{as:classical},
\begin{align*}
\sigma_{F_j}^2 \equiv & \sigma_j^2 \leq \sup_{f \in F_j} \|{\cal L}_f\|_{L_2}^2 = \sup_{f \in F_j} \E \left(\ell(f(X)-Y)-\ell(f^*(X)-Y)\right)^2
\\
\leq & L^2 B \sup_{f \in F_j} \E {\cal L}_f \leq L^2 B \cdot 2^{j}r^2.
\end{align*}
Observe that if $u_j = 2^{j-2}r^2$ and
$$
\sup_{f \in F_j} \left|\frac{1}{N}\sum_{i=1}^N {\cal L}_f(X_i,Y_i)-\E {\cal L}_f \right| \leq u_j,
$$
then
$$
\sup_{f \in F_j} \left|\frac{P_N{\cal L}_f}{\E {\cal L}_f} - 1 \right| \leq \frac{u_j}{\E {\cal L}_f} \leq \frac{1}{2},
$$
(because in $F_j$, $\E{\cal L}_f \geq 2u_j$) which is the ratio estimate one requires for the proof of Theorem \ref{thm:BBM}.

Applying the Gin\'{e}-Zinn symmetrization theorem,
\begin{align} \label{eq:class-eq-1}
& Pr\left(\sup_{f \in F_j} \left|\frac{1}{N}\sum_{i=1}^N {\cal L}_f(X_i,Y_i)-\E {\cal L}_f \right| \geq u_j \right) \nonumber
\\
\leq & 2 Pr\left(\sup_{f \in F_j} \left|\frac{1}{N}\sum_{i=1}^N \eps_i{\cal L}_f(X_i,Y_i)\right| \geq \frac{u_j}{4} \right),
\end{align}
provided that $u_j \geq 4N^{-1/2}\sigma_j$. Since $\sigma_j^2 \leq L^2B2^j r^2$ one may choose
\begin{equation} \label{eq:u-j}
2^{j-2}r^2=u_j \geq 4L\sqrt{B}2^{j/2}r/\sqrt{N}
\end{equation}
for the symmetrization argument to be valid, and which is a `legal' choice if $r \gtrsim \sqrt{B}L/\sqrt{N}$ as has been assumed.

Given $(X_i,Y_i)_{i=1}^N$, recall that $\xi_i=f^*(X_i)-Y_i$ and put $\phi_i(z)=\ell(z-\xi_i)-\ell(\xi_i)$.  Observe that
$$
{\cal L}_f (X_i,Y_i) = \ell \left((f-f^*)(X_i)+\xi_i\right)-\ell\left(\xi_i\right),
$$
and thus,
$$
{\cal L}_f(X_i,Y_i)=\phi_i\left((f-f^*)(X_i)\right).
$$
Clearly, $\phi_i(0)=0$ and $\|\phi\|_{\rm lip} \leq L$. The contraction theorem for Bernoulli processes \cite{LT:91} shows that for every fixed $(X_i,Y_i)_{i=1}^N$, one has
\begin{align} \label{eq:class-eq-2}
& Pr_\eps \left(\sup_{f \in F_j} \left|\frac{1}{N}\sum_{i=1}^N \eps_i{\cal L}_f(X_i,Y_i)\right| > \frac{u_j}{4} \right) \nonumber
\\
\leq & 2 Pr_\eps \left(\sup_{f \in F_j} \left|\frac{1}{N}\sum_{i=1}^N \eps_i (f-f^*)(X_i)\right| > \frac{u_j}{4L}\right).
\end{align}
By the Bernstein condition, it follows that for every $f \in F$, $\|f-f^*\|_{L_2}^2 \leq B \E {\cal L}_f$, and therefore $F_j \subset f^*+(H_{f^*} \cap \sqrt{B} 2^{j/2}r D)$. Hence, combining \eqref{eq:class-eq-1} and \eqref{eq:class-eq-2},
\begin{align*}
& Pr\left(\sup_{f \in F_j} \left|\frac{1}{N}\sum_{i=1}^N {\cal L}_f(X_i,Y_i)-\E {\cal L}_f \right| \geq u_j \right)
\\
\leq & 4 Pr \left(\sup_{h \in H_f^* \cap \sqrt{B} 2^{j/2}r D} \left|\frac{1}{N}\sum_{i=1}^N \eps_i h(X_i)\right| > \frac{u_j}{4L}\right).
\end{align*}

Applying Theorem \ref{thm:talagrand-conc} to the class $H_j = H_{f^*} \cap \sqrt{B} 2^{j/2}r D$, one has that with probability at least $1-2\exp(-x_j)$,
\begin{align*}
\sup_{h \in H_j} \left|\frac{1}{N}\sum_{i=1}^N \eps_i h(X_i)\right| \leq & C \left(\E\sup_{h \in H_j} \left|\frac{1}{N}\sum_{i=1}^N \eps_i h(X_i) \right| + r \sqrt{\frac{B 2^{j} x_j}{N}}+ b\frac{x_j}{N} \right)
\\
\leq & C\left(r^2 \gamma B 2^j + r \sqrt{\frac{B 2^{j} x_j}{N}}+ b\frac{x_j}{N} \right) \leq \frac{2^{j-2}r^2}{4L} = \frac{u_j}{4L},
\end{align*}

provided that $\gamma \lesssim 1/L B$ and $x_j \lesssim Nr^22^j \min\{1/Lb,1/LB\}$. Hence, by the union bound, with probability at least
\begin{align*}
& 1-2\sum_{j \geq 0} \exp\left(-c_0Nr^22^j \min\left\{(Lb)^{-1},(LB)^{-1}\right\}\right)
\\
\geq & 1-2\exp\left(-c_1\min\left\{(Lb)^{-1},(LB)^{-1}\right\}Nr^2\right),
\end{align*}
for every $j \geq 0$,
$$
\sup_{h \in H_f^* \cap \sqrt{B} 2^{j/2}r D} \left|\frac{1}{N}\sum_{i=1}^N \eps_i h(X_i)\right| < \frac{u_j}{4L}.
$$
Thus, for $r=\max\{\bar{k}_N(\gamma),L\sqrt{B}/\sqrt{N}\}$, one has
$$
Pr\left(\sup_{f \in F : \E {\cal L}_f \geq r} \left|\frac{P_N{\cal L}_f}{\E {\cal L}_f} - 1 \right| \leq  \frac{1}{2}\right) \geq 1-8\exp\left(-c_1Nr^2\min\left\{(Lb)^{-1},(LB)^{-1}\right\}\right),
$$
implying that with probability at least $1-\delta$,
$$
{\cal E}_p \leq c_2 \max\left\{ \left({\bar k}_N\left(\frac{c_3}{LB}\right)\right)^2,\frac{L^2B}{N}, \frac{\log(1/\delta)}{N}\cdot L\max\{b,B\}\right\}.
$$
\endproof

\section{The persistence problem via Theorem \ref{thm:BBM}} \label{app:persistence}
For every $n$, let $\mu_n$ be a measure on $\R^n$, set $T_{r,n}$ to be an increasing hierarchy of subsets of $\R^n$ and define $F_{r,n}=\left\{\inr{t,\cdot} : t \in T_{r,n}\right\}$ to be the classes of linear functionals associated with $T_{r,n}$.

Given a target $Y$ taken from a reasonable family of targets, consider the prediction and estimation problems in $F_{r,n}$ with $X \sim \mu_n$ and with respect to the squared loss.

The goal is to identify the largest `radius' $r(N)$ and dimension $n(N)$, as a function of the sample size $N$, for which ${\cal E}_p$ and ${\cal E}_e$ still tend to zero as $N$ tends to infinity.

Note that the solution of the persistence problem depends on obtaining sharp estimates on ${\cal E}_p$ and ${\cal E}_e$ for each one of the classes $F_{r,n}$ as a function of the radius $r$ and of the dimension $n$.

One hierarchy that has been studied extensively in the context of persistence, possibly because of its connections with sparse recovery problems, is
$$
T_{r,n}=\left\{x \in \R^n : \sum_{i=1}^n |x_i| \leq r\right\} \equiv r B_1^n,
$$
which are multiples of the unit ball in $\ell_1^n$.

Let $\mu_n$ be the uniform measure on $\{-1,1\}^n$ (i.e., $X=(\eps_1,...,\eps_n)$ for independent, symmetric $\{-1,1\}$-valued random variables). Fix $t_0 \in T_{r,n}$ and $\sigma>0$, let $\eps_{n+1}$ be a symmetric $\{-1,1\}$-valued random variable that is independent of $X$ and set $Y=\inr{t_0,\cdot}+\sigma \eps_{n+1}$.

To see how this framework fits Theorem \ref{thm:BBM}, observe that $f^*(X)=\inr{t_0,X}$ and that
$$
\E {\cal L}_{f_t} = \E\inr{t-t_0,X}^2=\|t-t_0\|_{\ell_2^n}^2,
$$
implying that $B=1$. Also, since $\mu$ is supported in $\{-1,1\}^n$, it follows that for every $t \in \R^n$,
$$
\|\inr{t,\cdot}\|_{L_\infty} = \max_{x \in \{-1,1\}^n} \inr{x,t} = \sum_{i=1}^n |t_i| =\|t\|_{\ell_1^n}.
$$

Thus, $\sup_{t \in rB_1^n} \|\inr{r,\cdot}\|_{L_\infty} = r$ and $\|Y\|_{L_\infty} = \|t_0\|_{\ell_1^n}+\sigma$.

Set
$$
\rho_N=
\begin{cases}
\frac{r^2}{\sqrt{N}} \sqrt{\log \left(\frac{2c_1 n}{\sqrt{N}}\right)} & \mbox{if} \ \ N \leq c_1 n^2
\\
\\
\frac{r^2n}{N} & \mbox{if} \ \ N >c_1 n^2.
\end{cases}
$$
The outcome of Theorem \ref{thm:BBM} is that with probability at least $1-2\exp(- c_2N \rho_N/r^2)$,
$$
{\cal E}_p \leq c_3\rho_N.
$$
However, the optimal rate for this problem (see, for example \cite{LM13} and \cite{Shahar-COLT}) is given by the following. Let
\begin{equation*}
v_1=
\begin{cases}
\frac{r^2}{N} \log\left(\frac{2c_1n}{N}\right) &  \mbox{if} \ \ N \leq c_1 n,
\\
\\
0 & \mbox{if} \ \ N > c_1 n
\end{cases}
\end{equation*}
and
\begin{equation*}
v_2 =
\begin{cases}
\frac{r\sigma}{\sqrt{N}} \sqrt{\log\left(\frac{2c_2n\sigma}{\sqrt{N}r}\right)} & \mbox{if } \ \  N \leq c_2n^2 \sigma^2/r^2
\\
\\
\frac{\sigma^2 n}{N} & \mbox{if} \ \ N >c_2n^2 \sigma^2/r^2.
\end{cases}
\end{equation*}
Then with probability at least $1-2\exp\left(-c_3N \min\{v_2,1\}\right)$,
$$
{\cal E}_p \leq c_4\max\left\{ v_1,v_2\right\}.
$$
The two estimate are a clear indication that Theorem \ref{thm:BBM} is not only restricted in its scope, it is also suboptimal within it, as it scales incorrectly with the `radius' $r$ (which corresponds to the $L_\infty$ bound on class members) and with the noise level $\sigma$.

\end{document}